\documentclass[twocolumn,3p,times]{elsarticle}
\usepackage[utf8]{inputenc}
\usepackage[T1]{fontenc}

\usepackage{natbib}\let\cite\citep
\usepackage{latexsym}
\usepackage{soul,color}

\usepackage{graphicx}
\usepackage{multirow}
\usepackage{amsmath}

\usepackage{epsdice} 
\usepackage{algorithm,algpseudocode} 
\usepackage{cleveref}

\newcommand{\dataset}{WikiDes}

\usepackage{xurl} 

\newcommand{\insplit}{topic-exclusive split}
\newcommand{\exsplit}{topic-independent split}

\newcommand{\Insplit}{Topic-exclusive split}
\newcommand{\Exsplit}{Topic-independent split}

\usepackage{pbox}

\usepackage{amssymb}

\journal{Information Fusion}

\begin{document}

\begin{frontmatter}



\title{\dataset{}: A Wikipedia-Based Dataset for Generating Short Descriptions from~Paragraphs}



\author[aff1]{Hoang Thang Ta~}
\ead{tahoangthang@gmail.com}
\author[aff2]{Abu~Bakar~Siddiqur~Rahman}
\ead{abubakarsiddiqurra@unomaha.edu}
\author[aff3]{Navonil Majumder}
\ead{navonil\_majumder@sutd.edu.sg}
\author[aff4]{Amir Hussain}
\ead{A.Hussain@napier.ac.uk}
\author[aff2]{Lotfollah~Najjar}
\ead{lnajjar@unomaha.edu}
\author[aff5]{Newton~Howard}
\ead{newton.howard@nds.ox.ac.uk}
\author[aff3]{Soujanya Poria}
\ead{sporia@sutd.edu.sg}
\author[aff1]{Alexander Gelbukh}
\ead{gelbukh@cic.ipn.mx}

\address[aff1]{Centro de Investigaci\'on en Computaci\'on (CIC), Instituto Polit\'ecnico Nacional (IPN), Mexico}
\address[aff2]{College of Information Science and Technology, University of Nebraska Omaha, USA}
\address[aff3]{ISTD, Singapore University of Technology and Design, Singapore}
\address[aff4]{Edinburgh Napier University, UK}
\address[aff5]{University of Oxford, UK}

\begin{abstract}

As free online encyclopedias with massive volumes of content, Wikipedia and Wikidata are key to many Natural Language Processing (NLP) tasks, such as information retrieval, knowledge base building, machine translation, text classification, and text summarization. In this paper, we introduce \dataset{}, a novel dataset to generate short descriptions of Wikipedia articles for the problem of text summarization. The dataset consists of over 80k English samples on 6987 topics.
We set up a two-phase summarization method --- description generation (Phase I) and candidate ranking (Phase II) --- as a strong approach that relies on transfer and contrastive learning. For description generation, T5 and BART show their superiority compared to other small-scale pre-trained models. By applying contrastive learning with the diverse input from beam search, the metric fusion-based ranking models outperform the direct description generation models significantly up to $\approx$ 22 ROUGE in \insplit{} and \exsplit{}. Furthermore, the outcome descriptions in Phase II are supported by human evaluation in over 45.33\% chosen compared to 23.66\% in Phase I against the gold descriptions. In the aspect of sentiment analysis, the generated descriptions cannot effectively capture all sentiment polarities from paragraphs while doing this task better from the gold descriptions. The automatic generation of new descriptions reduces the human efforts in creating them and enriches Wikidata-based knowledge graphs. Our paper shows a practical impact on Wikipedia and Wikidata since there are thousands of missing descriptions. Finally, we expect \dataset{} to be a useful dataset for related works in capturing salient information from short paragraphs. The curated dataset is publicly available at: \url{https://github.com/declare-lab/WikiDes}.

\end{abstract}




\begin{keyword}
Text summarization \sep contrastive learning \sep sentiment analysis \sep metric fusion \sep Wikipedia \sep Wikidata


\end{keyword}

\end{frontmatter}


\section{Introduction}

Text summarization is the task of producing a summary that keeps salient information of a certain document. The task can be monolingual or cross-lingual~\cite{Cross-langsumm-Polib}. The  monolingual task has been addressed for languages other than English~\cite{Kaz-Cys}. In dataset building, the document and the summary are aligned as a pair, having various lengths depending on the purpose of use. Free online platforms such as Wikipedia and Wikidata provide massive-scale and diverse content to assemble information for summarization tasks. A Wikidata item contains a short descriptive phrase to distinguish between items with the same or similar labels. In the mobile version, these descriptions appear on the top of Wikipedia articles, helping users to perceive the topic of articles they want to read~\cite{Sakota2022Descartes}. Since being sister projects, Wikidata and Wikipedia have connected each other by interwiki links which store on Wikidata. We observe a high correlation between Wikidata descriptions and Wikipedia articles, especially in the first paragraphs. Therefore, our objective is to construct a novel dataset named \dataset{} for generating short descriptions as summaries from the first paragraphs as documents. \dataset{} is a monolingual dataset with over 80k English samples with 6987 instances as topics, extracted data from both Wikipedia and Wikidata. 


With the rapid development of Wikipedia and Wikidata in recent years, the editor community has been overloaded with contributing new information adapting to user requirements, and patrolling the massive content daily. Hence, the application of NLP and deep learning is key to solving these problems effectively. In this paper, we propose a summarization approach trained on \dataset{} that generates missing descriptions in thousands of Wikidata items, which reduces human efforts and boosts content development faster. The summarizer is responsible for creating descriptions while humans toward a role in patrolling the text quality instead of starting everything from the beginning. Our work can be scalable to multilingualism, which takes a more positive impact on user experiences in searching for articles by short descriptions in many Wikimedia projects.

Some authors mentioned different types of summarization, which depend on user requirements in their survey works 
about text summarization systems~\cite{gholamrezazadeh2009comprehensive,el2021automatic}. If classifying by summary content, there are indicative summarization and informative summarization. Indicative summarization systems retrieve about 5 to 10 percent of content as the main idea from the document. This kind of summarization has the role of encouraging users to continue reading the document because the summary only provides brief information about the subject of the document. Meanwhile, informative summarization systems offer a brief version, which can replace the main document~\cite{gholamrezazadeh2009comprehensive}. In our view, description generation is an indicative summarization task where descriptions are short text pieces and just enough to let users know about the article topic. For example, \textit{"family"} is the description of article \textit{"The House of FitzJames"}. 

\citet{Sakota2022Descartes} conducted similar work to our paper on generating short descriptions from paragraphs and considered it a type of extreme summarization~\cite{narayan2018don}. They used Wikipedia's first paragraphs, Wikidata descriptions, and Wikidata instances as the input for the summarization. Their multilingual dataset is massive-scale when capturing content from Wikipedia and Wikidata in over 25 languages, which makes the model architecture bulky with a high training cost. Therefore, they chose to turn the custom attention mechanism instead fine-tune the encoder over languages in the training process. In contrast, we offer an available English dataset that uses Wikipedia's first paragraphs as the input and Wikidata descriptions as the output. Our work relies on the correlation between Wikipedia and Wikidata which the input is independent of the output. The main task is to generate summaries for a project from the content of another project. We apply two-phase summarization --- description generation and candidate ranking --- to improve the generation performance supported by contrastive learning and beam search. Thus, the generated descriptions capture more salient information from paragraphs, making our summarization task less "extreme". For example, \textit{"noble house founded by James FitzJames, 1st Duke of Berwick"} is the generated description of the article \textit{"The House of FitzJames"} instead of the gold description \textit{"family"}.


In short, we introduce \dataset{}, a novel dataset for description summarization, which creates short descriptions in Wikidata style from given paragraphs. The main contributions are as follows:
\begin{itemize}
    \item \textit{Dataset creation}: We provide an available dataset on GitHub\footnote{\url{https://github.com/declare-lab/WikiDes}} for research purposes related to Wikipedia summarization.
    \item \textit{Setting up a trending approach}: We apply two-phase summarization -- description generation and candidate ranking~\cite{liu2021simcls,zhong2020extractive} -- to improve the quality of the generated descriptions. In more detail, we deploy transfer learning from small-scale pre-trained models such as T5 and BART for description generation and contrastive learning for candidate ranking by fusing metrics, such as cosine similarity and ROUGE. 
    \item \textit{Sentiment correlation}: We measure the correlations of the generated descriptions versus the paragraphs and the gold descriptions by comparing their sentiment polarities on cumulative distribution and the Kolmogorov-Smirnov test.

\end{itemize}

The remainder of this paper is organized as follows: Section 2 describes related works to the summarization tasks, especially ones related to Wikipedia and Wikidata. Section 3 and Section 4 introduce how to create \dataset{} and perform several data analyses on it. We present our methods of description generation and sentiment measurement in Section 5 and use them for the experiments in Section 6 with human evaluation and error analysis. Finally, we give our conclusion and outline the  future work in Section 7.


\section{Related Works}

In this section, we present existing datasets and deep learning approaches for text summarization not only in wiki text but also in other domains. The works on integrating sentiment analysis into text summarization are also outlined by some prominent approaches. Besides, the evaluation metrics such as ROUGE which evaluate the quality of generated summaries over the overlap of semantic units or embeddings similarity are mentioned in \Cref{eval_metric}.



\subsection{Datasets Extracted from Wikipedia and Wikidata}

Monolingual and multilingual are two types of datasets that collect content as articles and knowledge graphs as RDF triples from Wikipedia and Wikidata. Monolingual datasets are usually in English because the most coverage and massive-scale content appear in English Wikipedia in a great collaborative effort from millions of editors. On another point, multilingual datasets capture the multilingualism strength of Wikipedia when this encyclopedia supports up to 327 language editions\footnote{\url{https://meta.wikimedia.org/wiki/List_of_Wikipedias}}.

Monolingual datasets are built from several popular languages, of which English and German are two typical examples. \citet{Field2020A_Generative} built a large dataset in English, including 14.4M articles with section titles and their content. The summarization task is to create a title from a given section's content. \citet{Frefel2020kSummarization} collected a large corpus of 240,000 texts in German Wikipedia. For each article, they considered the first paragraph as a summary and the rest of the content as the document.
\citet{Zhu2022Queryfocused} introduced WIKIREF, a large query-focused summarization dataset collected from Wikipedia articles with more than 280,000 examples. 
This research benefits from the information synthesis process of Wikipedia editors in writing articles.

Multi-document summarization (MDS) exists in some monolingual datasets. \citet{Zopf2016hMDS} proposed \textit{\textbf{h}}MDS, a new, heterogeneous, and multi-genre corpus for MDS. 
Later, the same first author upgraded this work to auto$-$\textit{\textbf{h}}MDS, a multilingual multi-document summarization (MMS) dataset~\cite{zopf2018auto}. \citet{Antognini2020kGeameWikiSum} built GameWikiSum, an MDS dataset in the game domain with 14,652 samples. The dataset has video game reviews taken from online platforms such as Play Station or Xbox as documents and gameplay sections of Wikipedia articles as summaries. Another MDS dataset extracted content from Wikipedia Current Events Portal (WCEP) was collected by \citet{Ghalandari2020A_LargeScale} to provide summaries for news events.

To convert a dataset from monolingual to multilingual, we can use machine translation systems to create pseudo-cross-lingual summarization data. However, this method may contain noises from the translation process so there is better to create cross-lingual datasets without translation. \citet{fatima2021ANovel} collected a cross-lingual dataset from Spektrum der Wissenschaft and Wikipedia's Science portals (WSP) with 51,312 English and German scientific articles.
\citet{Perez-Beltrachini2021XWiki} selected Wikipedia body paragraphs and leading paragraphs as document-summary pairs and their cross-lingual dataset XWikis contains 12 languages. Later, \citet{Pavel2022WikiMulti} built WikiMulti from Wikipedia good articles. For each article, the first paragraph is a summary, and the remaining content is a document. WikiMulti contains 22,061 unique English articles and 9,639 articles on average in other languages that align with English articles.


\subsection{Other Summarization Datasets}


Some early notable datasets are Document Understanding Conference (DUC) and Text Analysis Conference (TAC), sponsored by NIST for the summarization task on a small set of documents. DUC had a series from 2001-2007, then became a Summarization track of TAC in 2008. They both focused on generic and question summaries of English newspapers and newswire articles. DUC contains two evaluation methods: a baseline by an automatic system in NIST and human evaluation by the linguistic quality and conciseness~\cite{Over2007DUC}. As a pioneer of the guided summarization task, TAC 2010 required generating a 100-word summary for a given topic from a set of 10 Newswire articles~\cite{Genest2012Guided}.

In the domain of news content, Gigaword, New York Times (NYT), CNN/Daily\-Mail newspapers (CNNDM or CNN/DM), and Newsroom are several examples of popular datasets, created by collecting hundreds to millions of news articles from publishers. Gigaword is an available large-scale corpus of English news with nearly 10 million documents from seven news outlets. 
NYT has over 65,0000 articles-summary pairs~\cite{sandhaus2008new} while CNNDM is more popular and contains 93K articles from CNN (Cable News Network) and 220K articles from DailyMail~\cite{Hermann2015CNNDailMail}. Newsroom is another large-scale dataset in the news domain, with 1.3 million summaries collected from 38 news publishers. The content has various writing styles due to extracting the source texts from diverse sources such as social media and articles on the Internet~\cite{grusky2018newsroom}. Different from mentioned datasets, Multi-News is the first news dataset that relied on the MDS of news articles.~\cite{fabbri2019multi}. From a set of new events, the main task is to generate a well-organized summary, which can cover all events comprehensively and simultaneously avoid redundancy. XSum is similar to our dataset, \dataset{} in terms of extreme summarization when generating a short, one-sentence news summary that just answers the question \textit{“What is the article about?”}~\cite{narayan2018don}. In our case, we generate a short description that can represent a given paragraph by taking the salient information.

News datasets usually contain typical writing styles of journalists with importance in the first paragraph. However, Wiki datasets bring a new wind of diversity with writing styles from ordinary people. WikiHow is a large corpus of more than 230,000 article and summary pairs. Only some common features of n-grams exist between the source articles and the reference summaries in WikiHow. The more unique n-grams between the source articles and the reference summaries, the more it is challenging to generate a quality summary compared to the reference summary~\cite{koupaee2018wikihow}. WikiSum has the same knowledge base as WikiHow but uses simple English in documents. Summaries are coherent paragraphs as tips written by the document authors in a friendly manner. Therefore, its content is highly readable and easily comprehensive for readers~\cite{cohen2021wikisum}. 

In the academic domain, Arxiv and PubMed are summarization datasets with coherent paragraph summaries in scientific writing styles~\cite{Cohan2018Dis}, and the document length is significantly longer than the one in the news domain. The content of arXiv and PubMed was retrieved from scientific papers on arXiv.org and PubMed.com. Another dataset, BIGPATENT contains 1.3 million records of U.S. patent documents and human-written abstractive summaries~\cite{Sharma2019BIGPATENT}. Besides Multi-News and WikiSum, Multi-XScience is also a large-scale MDS dataset collected from scientific articles~\cite{lu2020multi}. It brings a challenge for the MDS task: given a paper, generate a related-work section from the paper abstract and the abstracts of reference pieces. 

\subsection{Deep Learning Approaches for Text Summarization}

As a traditional task in NLP, text summarization has a long development history with many methods or approaches invented to address the summarization issues. Some prominent approaches can be named as statistical-based approaches, topic-based approaches (LSA~\cite{steinberger2004using}, topic themes~\cite{harabagiu2005topic}), and graph-based approaches (LexRank~\cite{erkan2004lexrank}, TextRank~\cite{mihalcea2004textrank}, Opinosis~\cite{ganesan2010opinosis}, GraphSum~\cite{baralis2013graphsum}), and approaches based on machine learning~\cite{gambhir2017recent}. In this section, we focus on deep learning approaches due to their effectiveness recently in generating quality summaries.

Many summarization tasks apply encoder-decoder architectures by different deep learning methods. \citet{Rush2015Neural} and \citet{Nallapati2016Abs} used an attention-based encoder-decoder model to create summaries in abstractive summarization. A neural sequence-to-sequence model sometimes reproduces incorrect factual details and repeats text pieces. To solve these problems, \citet{See2017GetToThePoint} proposed a pointer-generator network (PGN) to point to copy words from the source documents for supporting correct information reproduction and avoid repetition when tracking the sentences covered in the document. In another work, \citet{Nallapati2018Summaerrunner} proposed a two-layer bidirectional Gated Recurrent Unit (GRU) recurrent neural network-based (RNN) classifier to generate the extractive summary based on the content richness of each sentence and saliency based on the overall document. 
\citet{narayan2018don} built a topic-conditioned neural model based on convolutional neural networks (CNN). Compared to RNNs, convolution layers hold long-range dependencies between words better and allow doing the inference, abstraction, and paraphrasing at a document level. 


Recently, \textit{transformers} has emerged as a library to construct high-capacity models with transformer architecture and apply pretraining effectively for diverse NLP tasks, including text summarization~\cite{wolf2020transformers}. This library allows researchers can inherit many pre-trained models such as BERT~\cite{liu2019fine,Ma2022T-BERTSUM}, BART~\cite{lewis2019bart}, T5~\cite{raffel2020exploring}, and PEGASUS~\cite{Zhang2020Pegasus} to extend training on text summarization datasets not only for academics but also for industrial sectors. Instead of taking time to train datasets in the beginning, some works used transfer learning from these pre-trained models to their downstream tasks~\cite{laskar2020query,alomari2022deep}.

Using multilingual BART, \citet{Sakota2022Descartes} did a similar task to our work on generating short descriptions from the first paragraphs of Wikipedia articles, providing a quick insight into a Wikipedia article. They deployed Descartes (\underline{Description} of \underline
{art}icl\underline{es}), a model built based on a pre-trained multilingual BART model with data in 25 languages. In multilingual content generation, a created summary is usually in the same language as the document . In this case, helpful texts in other languages can not utilize. Authors thus leveraged the multilingualism of Wikipedia so that any summarizer can generate a description in a language without requiring the source document in that language~\cite{Sakota2022Descartes}. 
However, this method creates a bulky generation architecture and takes more natural noises from Wikipedia and Wikidata. Due to the openness of Wikimedia projects~\cite{javanmardi2009user}, the community sometimes can not prevent vandalism types on contribution content in many language editions. Hence, the more content used in more languages, the more noise the dataset may have.

Contrastive learning is a popular technique to maximize similarities of feature representation of the same images and minimize those in the different images. Based on a similar sense, \citet{Xu2022TSeqContrastive} used contrastive learning in summarization tasks by maximizing the similarities of the same semantic meaning articles. \citet{liu2021simcls} also used BART and RoBERTa in two-phase summarization to generate and score candidate summaries which were generated from diverse beam search~\cite{vijayakumar2016diverse} to improve the qualities of output texts. The benefit of two-phase summarization mainly lies in the decoding strategy by beam search and other superior methods such as nucleus sampling~\cite{holtzman2019curious}, where diverse data creates more opportunities to search for an "ideal" candidate. In one of the most recent works, \citet{liu2022brio} followed a new training paradigm that assigns probabilities of candidate summaries concerning to their quality in contrastive learning. 
So far, some other summarization works have been done with contrastive learning in different methods~\cite{Shi2019DeepChannel,Junpeng2021TopicContrastive,Shuyang2021CLIF,Wang2021ContrastiveMultilingua,Liu2021CO2sum}. In addition, there are different approaches, including few-shot and zero-shot learning~\cite{goodwin2020flight}, reinforcement learning~\cite{alomari2022deep}, prompting~\cite{aghajanyan2021htlm}, prefix-tuning on massive-scale models (GTP-2)~\cite{li2021prefix}, hybrid approaches~\cite{zhang2019hie}, and others for various types of text summarization.

\subsection{Sentiment Analysis in Text Summarization}




An ideal and typical summary is able to capture the salient information from a given document~\cite{ghosal2022cice,hazarika2021conversational}. However, it may lack sentiment information of the document, which are key in datasets like IMDB movie reviews. For capturing sentiment, sentiment analysis is incorporated into text summarization models.

One of the integration approaches is to extract sentiment polarity (negative, neutral, and positive scores) in texts \cite{yadav2016text}, which allows us to turn sentiment analysis into a problem of text sentiment classification~\cite{ohana2009sentiment,dogan2019deep,shetty2015auto}. \citet{beineke2004exploring} introduced a method for sentiment summary generation, which discovers a key aspect of the author’s opinion in movie reviews on the Rotten Tomatoes website and applied Naive Bayes and regularized logistic regression to fit features extracted from good summaries in text classification. Another approach to combining sentiment analysis in text summarization is aspect-based sentiment summarization (ABSA), which contains two tasks, aspect identification with mention extraction and sentiment classification. \citet{wu2016aspect} mentioned a definition named Aspect-based Opinion Summarization (AOS) with the same tasks as ASBA. \citet{titov2008joint} presented a joint model of text and aspect ratings, which used a Multi-Aspect Sentiment model (MAS) to form representative aspects as topics and a set of sentiment predictors to demonstrate the correlations between a topic and a particular aspect. In another work, \citet{dhanush2016aspect} designed an RNN for extracting aspects with contexts in a sentence and a CNN for sentiment classification at the sentence level. 
Several works exploited other sentiment information in texts. For example, \citet{nishikawa2010optimizing} addressed informativeness and readability in restaurant reviews. They set up an algorithm to create a summary by choosing and arranging sentences by informativeness and readability scores. \citet{lerman2009sentiment} utilized the benefit of user preferences to construct a new summarizer on sentiment-based models.

In \dataset{}, a large number of texts are neutral due to the neutrality policy of Wikipedia, so we prefer to determine the uniformity of the sentiment polarity distribution between descriptions and paragraphs by cumulative distribution and the Kolmogorov-Smirnov test~\cite{massey1951kolmogorov} more than performing sentiment summarization approaches to capture other sentiment polarities. Besides the Kolmogorov-Smirnov test, some proper methods for measuring the distribution among multiple sets are the chi-squared test~\cite{plackett1983karl}, the Mann–Whitney U test~\cite{mcknight2010mann}, and the Fisher's z-test~\cite{lawley1938generalization}.

\section{Data Creation}

We rely on the APIs of Wikidata\footnote{\url{https://www.wikidata.org/w/api.php}} and English Wikipedia\footnote{\url{https://en.wikipedia.org/w/api.php}} to collect the data in JSON format. \Cref{algorithm:collect_data} is a simple algorithm for selecting samples randomly to avoid the bias of topic distribution in \dataset{}. The \texttt{is-a} or \texttt{instance-of} attribute of the Wikidata items is considered as the topic of the items. 
We eliminate the Wikidata items with following the topics because they do not have any corresponding articles in Wikipedia: 
\begin{itemize}
    \item  Scholarly article
    \item  Wikimedia disambiguation page
    \item  Wikinews article
\end{itemize} 


Given the number of samples $N$ needed to collect, a $while$ loop is used to extract output samples $S$. For each loop, an $id$ is randomly generated from 1 to 99 million\footnote{ \url{https://www.wikidata.org/wiki/Wikidata:Statistics}}. If the $id$ does not appear in the scanned list $K$, then continue to extract information from Wikidata and Wikipedia to build a sample. Next, we validate the sample, e.g., lengths of paragraphs and descriptions. If it is a good sample, then save it to $S$. Otherwise, the algorithm continues the $while$ loop until the number of samples $S$ equals the desired given number $N$.

\begin{figure*}[ht]
\centering
\includegraphics[width=\textwidth]{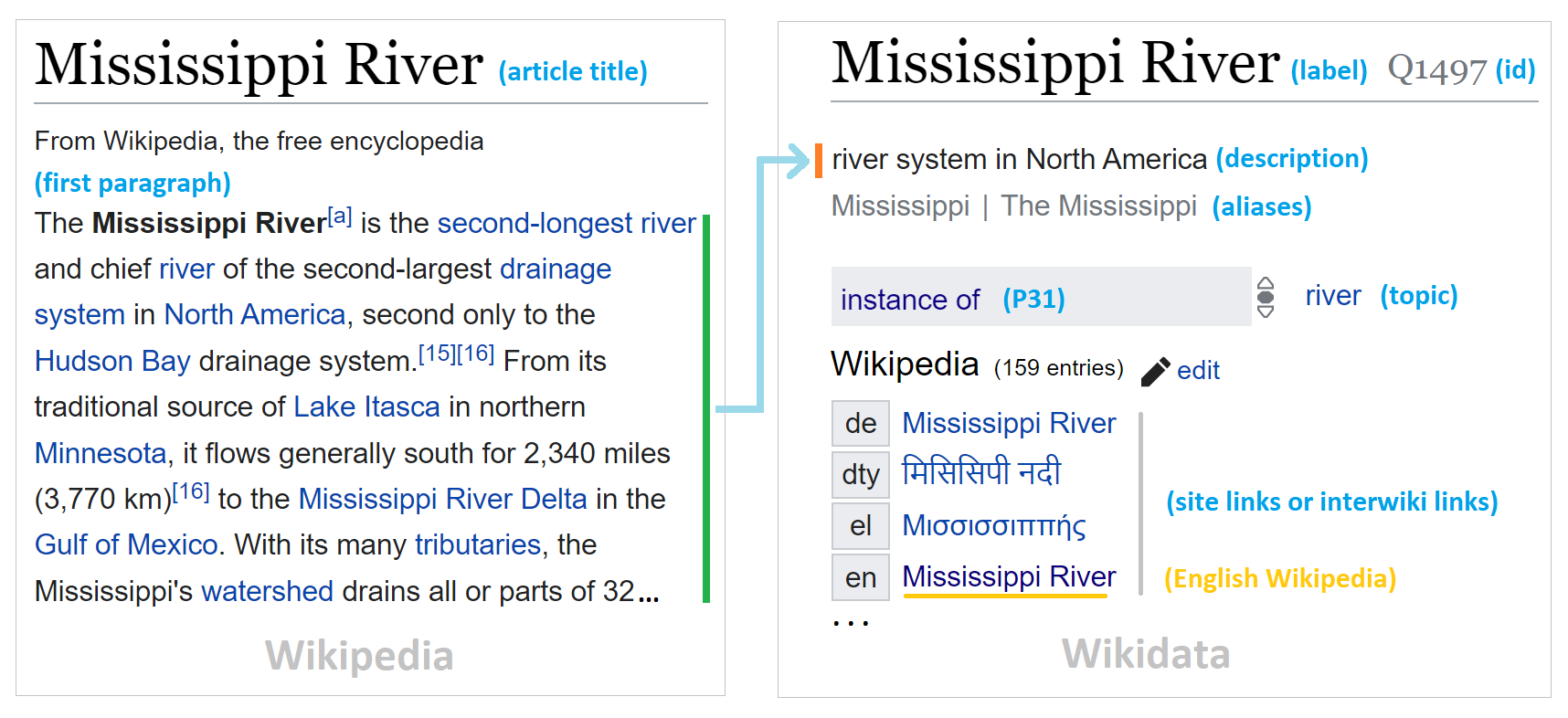}
\caption{A random sample in \dataset{}: The first paragraph (shown in a green bar) of the Wikipedia article is used to infer the description (shown in an orange bar) of the corresponding item \texttt{Q1497} in Wikidata.}
\label{fig:random_sample}
\end{figure*}

\begin{algorithm}
    \caption{Collect samples from Wikipedia and Wikidata} \label{algorithm:collect_data}
        \textbf{Input:} \textit{N}
            
        \textbf{Output:}
        \textit{S}
        \begin{algorithmic}
            
        \State $S, K \gets [], []$
            
        \While{$len(S)\le N$}
            \State $id\gets random(1, 99000000)$
                
            \If{$id \notin K $}
                \State $description, instances,..., en\_site\_link  
                \gets extractWikidata(id)$
                
                 \State $title, first\_para, first\_sen  \gets extractWikipedia(en\_site\_link)$
                 
                 \State $ description \gets preProcessing(description)$
                 
                 \State $first\_para \gets preProcessing(first\_para)$
                 
                 \State $first\_sen \gets extractFirstSentence(first\_para)$
                 
                 \State $sample \gets concateTuple(description, instances,...,first\_sen)$
                 
                 \If{($sample$ is good)}
                 
                 \State add $sample$ to $\textit{S}$

                 \EndIf
                 \State add $id$ to \textit{K}
            \EndIf
        \EndWhile
                            
    \end{algorithmic}
\end{algorithm}

For each item in Wikidata, the crawler extracts a label, a short description, instances (\texttt{P31}), and a site link or interwiki link. This link leads to an article in English Wikipedia, where the crawler can gather the first paragraph. We apply a few pre-processing steps to paragraphs and descriptions to remove special symbols and redundant spaces, which are assumed not to contribute effectively to the model performance in the description generation process. Furthermore, we discard samples with empty descriptions or paragraphs having fewer than 10 tokens.

Redirect articles are also discarded because they link to their representative articles, whose first paragraphs may not necessarily stand for the subjects of the redirect articles. For example, \texttt{Dong Nguyen}, a game developer links to the main article named \texttt{.Gears}, which is his company and more famous. In this case, we can not use the first paragraph of article \texttt{.Gears} representing for \texttt{Dong Nguyen}. In other technical details, we use the package \texttt{NLTK punkt} to extract the first sentences from the first paragraphs. Due to using online APIs, we applied multithreading to our crawler with \texttt{max\_workers=8} to speed up the data collection process. In less than 72 hours, 81,418 English samples on 6,987 topics were collected to build the dataset for this paper.

\Cref{fig:random_sample} describes a typical sample, which contains two essential texts: the Wikidata description \textit{"river system in North America"} and the first paragraph of the corresponding Wikipedia article \textit{"The Mississippi River is the second-longest river and chief river of the second-largest drainage system in North America..."}. The task is to generate this description from the given paragraph. The first element of the instance list (\textit{"river"}) is the baseline candidate in the experiment.

\section{Dataset Analyses}
In this section, we perform some analyses on the collected dataset, which are text length, instance distribution, and token position distribution. Later, we compare \dataset{} to existing popular datasets by several features in the summarization task.

\subsection{Data Distribution}

\begin{figure*}[ht]
\centering
\includegraphics[width=\textwidth]{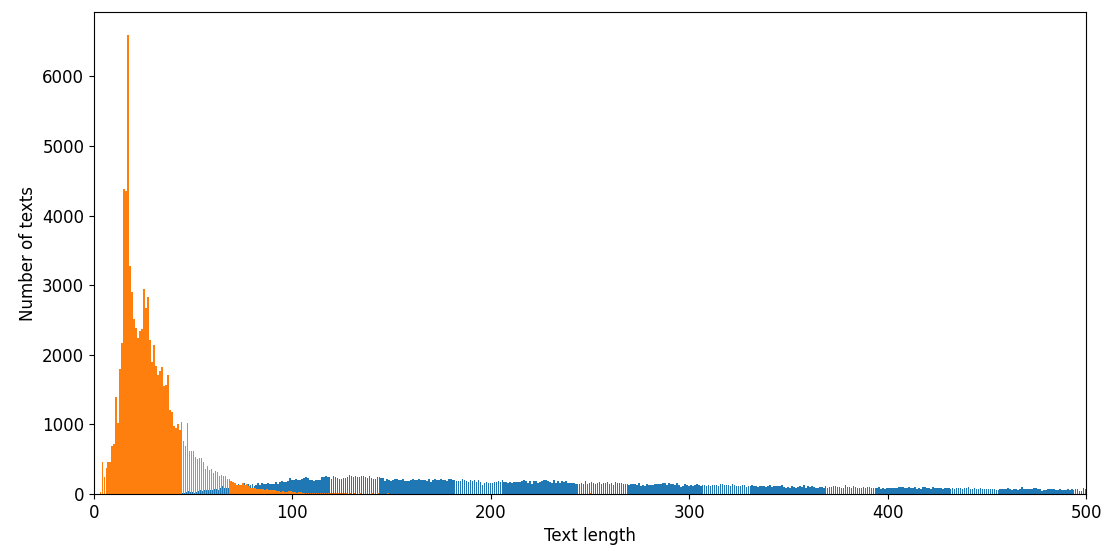}
\caption{The distribution of paragraphs (blue) and descriptions (orange) by their lengths, limited to 500 first lengths.}
\label{fig:text_length}
\end{figure*}

We count the lengths of paragraphs and descriptions as the number of words, without considering punctuations. \Cref{fig:text_length} shows the distribution of texts by lengths in the first 500 lengths. The average length of descriptions is 4.5 while the average length of paragraphs is 82.24. The descriptions are short with the majority of parts less than 100 words. Meanwhile, the paragraph distribution by length is more evenly compared to the descriptions.

\begin{figure}[ht]
\centering
\includegraphics[width=0.9\linewidth]{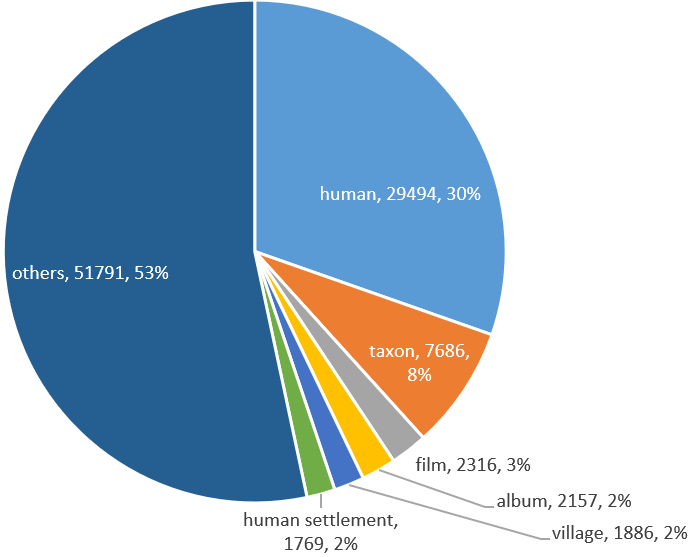}
\caption{The distribution of instances in the dataset.}
\label{fig:instance_distribution}
\end{figure}

Next, we check the instance distribution by samples in \dataset{} with 6987 instances. Wikidata instances (P31) belong to a complex hierarchy system of hypernyms and hyponyms, defined and built by its user community. In \dataset{}, some most popular instances by descending order are \texttt{human} (30\%), \texttt{taxon} (8\%), \texttt{film} (3\%), \texttt{album} (2\%), \texttt{village} (2\%), and \texttt{human settlement} (2\%), shown in \Cref{fig:instance_distribution}. The dataset also contains 3905 rare instances (55.88\%), appear in only 1 sample, such as \texttt{miniature war gaming} or \texttt{mask stone}. 

\subsection{Correlation Between Paragraphs and Descriptions}
\label{section:correlation_para_des}

Currently, most pre-trained models for the summarization task such as  BART and T5 support the maximum length of 1024 tokens. As shown in \Cref{fig:text_length}, the paragraphs are relatively short. Before the training process, we check the overlap rate between paragraphs and descriptions to know a proper maximum length of paragraphs for the training and reduce the capacity of input data. In the training set, we chop descriptions into text chunks with different token sizes (32, 64, 128, 256, 512, and 1024). The proper metric for this correlation is ROUGE-N-precision, counted by the number of overlap grams between descriptions and paragraphs over the number of grams in descriptions. \Cref{tab:correlation_paragraph_description} shows ROUGE-1, ROUGE-2, and ROUGE-L (or ROUGE-LCS, LCS refers to Long Common Subsequence) values between paragraphs and descriptions. Hence, 256 is an optimized length of paragraphs for the data training. ROUGE scores after 256 tokens have a little increment, which is less than 0.2 ROUGE while we have to feed into the model more than twice or fourth times of tokens.

\begin{table*}[ht]
\small
\centering
\caption{ROUGE scores between paragraphs and descriptions of the training set by different numbers of first tokens. The scores were calculated on 5000 random samples because it takes so much time to perform on the whole training set.}
\begin{tabular}{l|l|l|l}
\hline
number of tokens & R-1-precision & R-2-precision & R-L-precision \\
\hline
32 & 51.39 & 27.89 & 48.51 \\
\hline
64 & 60.99 & 33.31 & 57.24 \\
\hline
128 & 63.13 & 34.09 & 59.13 \\
\hline
\textbf{256} &  \textbf{63.72} & \textbf{34.24} & \textbf{59.70} \\
\hline
512 & 63.89 & 34.24 & 59.83 \\
\hline
1024 & 63.91 & 34.25 & 59.84 \\
\hline
\end{tabular}
\label{tab:correlation_paragraph_description}
\end{table*}

\begin{figure*}[ht]
\centering
\includegraphics[width=\textwidth]{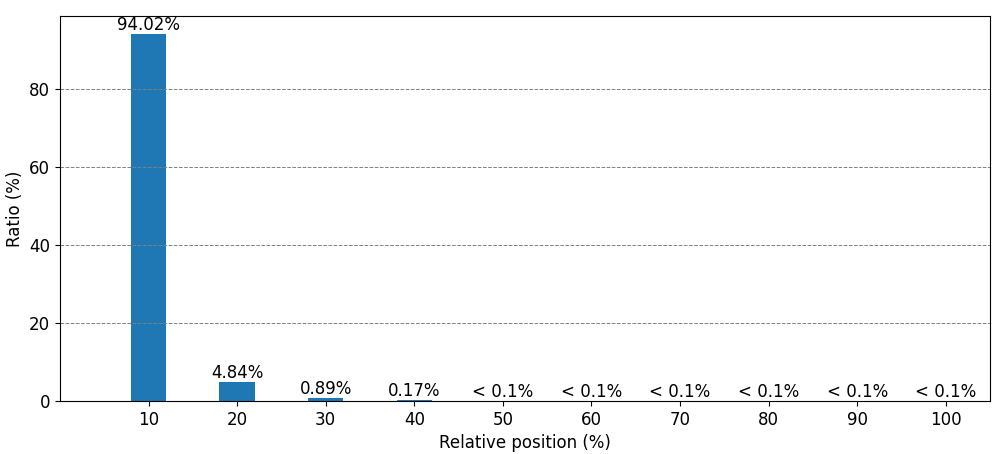}
\caption{The distribution of token positions of descriptions in paragraphs. Positions are converted to relative values, and their volumes are represented as ratios. The maximum position is 1024.}
\label{fig:token_distribution}
\end{figure*}

To support the optimized length of 256, we take another analysis on the token positions in paragraphs, displayed in \Cref{fig:token_distribution}. We encode paragraphs and descriptions by pre-trained model \texttt{facebook/bart-base} with the maximum length of 1024. Stop-words and punctuations are removed in descriptions; however, not from paragraphs. We see that most tokens appear 20\% of first positions, roughly 204.8 tokens. Therefore, this confirms the rationality of taking the maximum length of 256 tokens for the training process.

\subsection{Data Split}

In \Cref{fig:instance_distribution}, the instance distribution is used to split the collected dataset into training ($\approx$ 80\%), validation ($\approx$ 10\%), and test ($\approx$ 10\%) sets. Here, instances are topics, and we apply two split methods:
\begin{itemize}
    \item \textit{\Insplit{}}: We create a dictionary of topics by the number of samples and rank these topics from the most popular to rare. For example, we have \texttt{human} with 29494 samples, \texttt{taxon} with 7686 samples, \texttt{film} with 2316 samples, etc in \Cref{fig:instance_distribution}. The training set contains data with the top popular topics, while the other topics are allocated randomly to validation and test sets.
    \item \textit{\Exsplit{}}: In this split, we randomly put data into training, validation, and test sets without caring for topics.
\end{itemize}

\begin{table*}[ht]
\small
\centering
\caption{Topics in sets by topic-exclusive split and topic-independent split.}
\begin{tabular}{lll}
\hline
Data split & Training set & Validation and Test sets \\
\hline
Topic-exclusive & \pbox{6cm}{human, taxon, film, album, village, human settlement, family name, river, business, mountain, video game, chemical compound, organization, gene, radio station, road, high school, building, town, tennis event, fossil taxon, civil parish, airport, lake, city, island, language, military unit, school, political party, metro station, ..., reservoir, academic journal. \\ \textit{There are 156 topics in total.}} & \pbox{6cm}{software, cemetery, skyscraper, shopping center, free software, music genre, airbase, glacier, television channel, airline, sculpture, aircraft, valley, bridge, county seat, award ceremony, mosque, formation, manuscript, conflict, destroyer, publisher, poem, ..., college, protein, monastery, filmography, election, submarine, concept, trademark, toponym, atmosphere. \\ \textit{There are 6831 topics in total. These topics are allocated randomly into validation and test sets.}}  \\
\hline
Topic-independent & \pbox{6cm}{\textit{Topics are taken randomly from 6987 topics in every run time.}} & \pbox{6cm}{\textit{Topics are taken randomly from 6987 topics in every run time.}}\\
\hline
\end{tabular}

\label{tab:topic_distribution_by_set}
\end{table*}

\Insplit{} provides data on different topics to see how well the model can infer data on unseen topics. Otherwise, \exsplit{} offers the random distribution of data in sets, reflects the similarity of randomness in data creation, and avoids data bias. In specific, \Cref{tab:topic_distribution_by_set} shows the topics in training, validation, and test sets by \insplit{} and \exsplit{}. Meanwhile, \Cref{tab:other_datasets_comparison} represents the number of samples in different sets by these splits.

Due to a large number of instances with only one sample, we could not be able to split data by the same topics. These rare instances can be merged into their parents when they belong to a hierarchy system of hypernyms and hyponyms in Wikidata. However, this system is complex and overlaps due to a trade-off to user community~\cite{pellissier2020yago,shenoy2022study} with a freedom contribution in Wikidata. Therefore, it is sometimes difficult to choose the proper parents.

\subsection{Comparison with Other Summarization Datasets}

In this section, we compare our dataset with existing summarization datasets, especially those related to Wikipedia and Wikidata. Document length (doc. len.) and summary length (summ. len.) are counted by the average number of words appearing in texts. In our dataset, the number of words in paragraphs and descriptions is considered. We apply compression ratio (comp. ratio)~\cite{koupaee2018wikihow,wu2017topic}, calculated by the ratio between the average length of paragraphs and the average length of descriptions. This value offers a measurement of the task difficulty. The summarization task is more difficult with the higher value of compression ratio when the model has to deal with higher levels of abstraction and semantics. We counted only unique words without punctuations to build the vocabulary set in \dataset{}. The statistics of other datasets were extracted from their original papers or related papers in our best effort of searching. However, we self-calculated the compression ratio of these datasets based on the document length and the summary length that we got in hand.

\begin{table*}[ht]
\small
\centering
\caption{The comparison of our dataset with other existing datasets in the summarization task.}
\resizebox{\textwidth}{!}{
\begin{tabular}{lp{4cm}llll}
\hline
dataset & train/val/test & doc. len. & summ. len. & comp. ratio & vocab. size \\
\hline
\hline
arXiv$^{\beta}$~\cite{Cohan2018Dis} & 215K & 4938 & 220 & 22.44 & - \\
\hline
CNNDM$^{\alpha}$~\cite{cohen2021wikisum} & 287,113/13,368/11,490 & 789.9 & 55.6 & 14.20 & 717,951 \\
\hline
NEWSROOM$^{\alpha}$~\cite{grusky2018newsroom} & 1,321,995 &  658.6 & 26.7 & 24.66 & 6,925,712 \\
\hline
NYT$^{\alpha}$~\cite{cohen2021wikisum} & - & 795.9 & 44.9 & 17.72 & - \\
\hline
PubMed$^{\beta}$~\cite{Cohan2018Dis} & 133K & 3016 & 203 & 14.85 & - \\
\hline
Multi-News$^{\alpha}$~\cite{fabbri2019multi} & 44,972/5,622/5,622 & 2,103.49 &  263.66 & 7.97 & 666,515 \\
\hline
Multi-XScience$^{\beta}$~\cite{lu2020multi} & 30,369/5,066/5,093 & 778.08 & 116.44 & 6.68 & - \\
\hline
\textbf{\dataset{}}$^{\omega}$ & 65,772/7,820/7,827$^{\#}$ \newline 68,296/8,540/8,542$^{\epsdice{6}}$ & 82.24 & 4.50 & 18.27 & 354,946  \\
\hline
WikiHow$^{\omega}$~\cite{koupaee2018wikihow} & 230,843 & 579.8 & 62.1 & 9.33 & 556,461 \\
\hline
WikiSum$^{\omega}$~\cite{lu2020multi} & 15m/38k/38k & 1,334.2 & 139.4 & 9.57 & - \\
\hline
XSum$^{\alpha}$~\cite{narayan2018don} & 204,045/11,332/11,334 & 431.07 & 23.26 & 18.53 & - \\
\hline
\multicolumn{6}{p{15cm}}{$\alpha$: news, $\beta$: scientific document, $\omega$: wiki, \#: topic-exclusive, \epsdice{6}: topic-independent} \\
\hline
\end{tabular}
}
\label{tab:other_datasets_comparison}
\end{table*}

\Cref{tab:other_datasets_comparison} shows statistics of \dataset{} against other existing datasets, where \dataset{} has the lowest value of vocabulary size. In Wikimedia projects (Wikipedia, Wikidata, Wikibooks, etc.), content is freely contributed by millions of users from everywhere and complies with formal writing standards that offer easy readability for readers based on using popular words. Therefore, any data taken from Wikipedia and Wikidata is normally less diverse in terms of word usage. \dataset{} has a high compression ratio of 18.27, indicating that the summarization task on it is more difficult than some other datasets. There is a mismatch between set sizes of \insplit{} and \exsplit{} in \dataset{}. This is because we keep samples with empty instances in \exsplit{} while removing them in \insplit{}.


\section{Tasks and Models}

\subsection{Task Description}

There are 2 main tasks of this paper:
\begin{itemize}
    \item \textit{Two-phase summarization}: From a given paragraph $X$ as the input, the task is to generate the output as a set of candidate descriptions $\tilde{Y}_1$, $\tilde{Y}_2$, ..., $\tilde{Y}_n$, which is ranked by metric values in descending order. The list also includes the best description $\tilde{Y}_b$, which obtains the highest metric value.
    

    \item \textit{Sentiment correlations}: 
    Given two cumulative distributions $F_{1,N}(x)$ and $F_{2,M}(x)$ of two sets with the sample sizes $N$ and $M$ of the first set and second set correspondingly. Our task is to calculate a maximum distance $D$ between two sets and check this value against the rejection thresholds by some $\alpha$ levels of significance before concluding to accept or reject the null hypothesis, which is to answer the question: "Do two sets have the same distribution or not?".
    
    
\end{itemize}

\subsection{Method}

Previous approaches in text summarization use only one-phase summarization, in which the model learns how to generate summaries based on the probably next tokens in a sequence-to-sequence architecture. However, as a trending approach, two-phase summarization shows the effectiveness in improving the quality of output summaries in many works~\cite{liu2021simcls,zhong2020extractive,liu2021refsum,zhang2019pretraining}. The first phase is to train the summarization model before using it to infer a set of candidate summaries based on various decoding strategies~\cite{holtzman2019curious} such as beam search. The second phase applies contrastive learning~\cite{liu2021simcls,zhong2020extractive,Xu2022TSeqContrastive} to train a ranking model, which depends on the combination of semantic and lexical similarity of each element of the candidate set and the gold summary against the document to rank the output candidates by several popular metrics.

\begin{figure*}[ht]
\centering
\includegraphics[width=\textwidth]{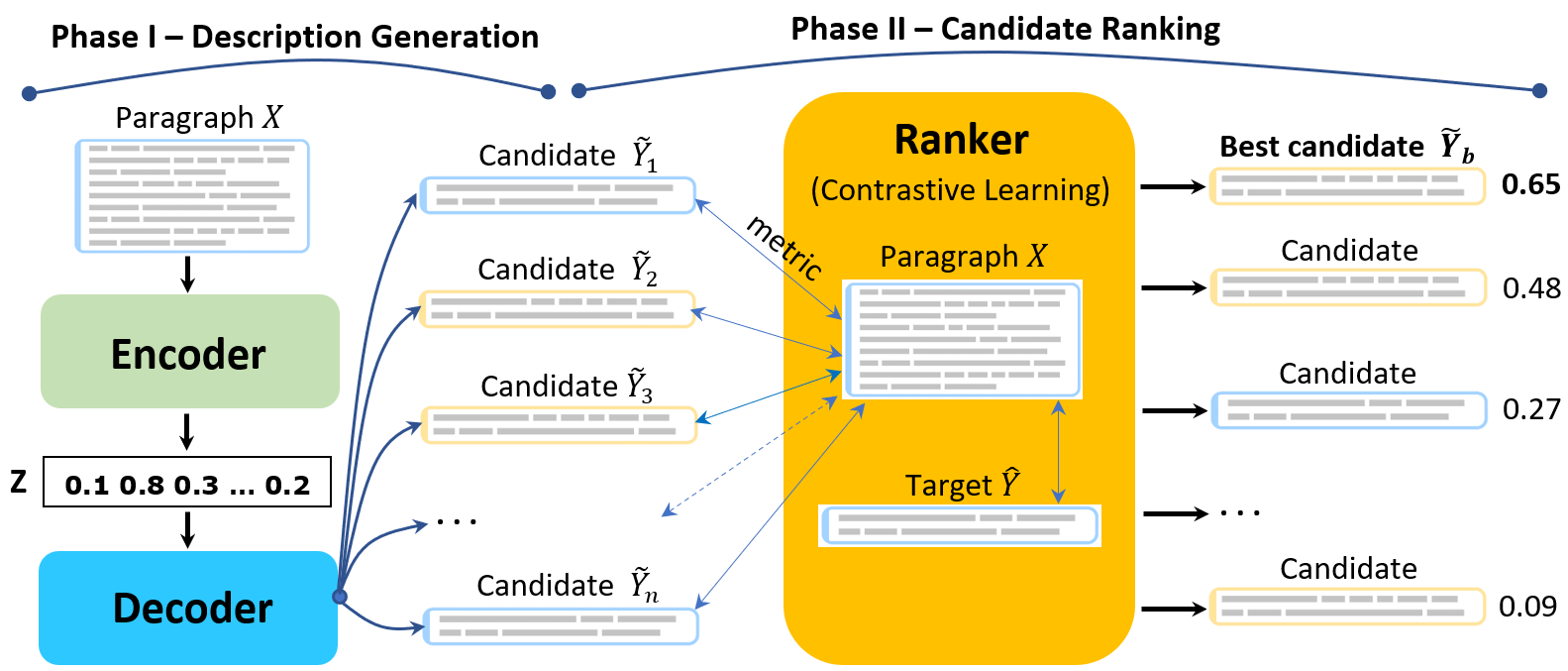}
\caption{The diagram of two-phase summarization used to generate the quality descriptions.}
\label{fig:two_phase_sum}
\end{figure*}

\Cref{fig:two_phase_sum} presents the diagram of two-phase summarization --- description generation and candidate ranking, which is used as the main method of this paper. In Phase I, the encoder transforms the given paragraph $X$ to the representation vector $Z$, then the decoder uses $Z$ to generate a set of candidate description $\tilde{Y}_1$, $\tilde{Y}_2$, ..., $\tilde{Y}_n$ based on beam search configurations. In Phase II, the ranker takes the candidate list and the gold (target) description $\hat{Y}$ for measuring the fusion of semantic and lexical similarity with the paragraph $X$. The output is a candidate list sorted descending by fused-metric values and the best candidate $\tilde{Y}_b$ will have the highest value. Finally, we investigate the uniformity of sentiment distribution on the output descriptions against the paragraphs and the gold descriptions by the cumulative test and the Kolmogorov-Smirnov test.



\subsubsection{Baseline Method for Description Generation}
\label{sec:baseline_method}

For each sample, there is a list of instances (\texttt{P31}) ranked by an order in Wikidata which expresses the \texttt{is-a} relations with the item. In that meaning, they are considered as topics of a certain description. We consider the first element of this list as the baseline description. For example, item \texttt{"Bugema University"} (Q4986155) contains two instances, \texttt{"university"} and \texttt{"church college"}. So we choose \texttt{"university"} as the baseline description.

\subsubsection{Description Generation Models}


The description generation is trained on small-scale pre-trained models with a Transformer architecture. Invented by \citet{vaswani2017attention}, a Transformer consists of three components -- an encoder, a decoder, and an attention mechanism -- mixed with recurrence and convolutions and applies a sequence-to-sequence structure for the conditional generation task.

Let define a source paragraph $X={x_1, x_2, x_3, ..., x_n}$ is a sequence of $n$ tokens and a target description $Y={y_1, y_2, y_3,...,y_m}$ is a sequence of $m$ tokens. The encoder simply converts $X$ into a representation sequence $Z = {z_1, z_2, z_3,...,z_n}$. The number of elements in $Z$ equals to the number of elements in $X$.

\begin{equation}
\begin{aligned}
Z = {z_1, z_2, z_3,..., z_n} = Encoder(x_1, x_2, x_3,...,x_n) \\
\end{aligned}
\label{eq:rep_sequence}
\end{equation}

Next, the decoder is responsible for generating the target description $Y$ from $Z$ in the equation $Y = Decoder(Z)$. Following the chain rule, the probability $p(Y|Z)$ that the decoder generates $Y$ from $Z$ is:
\begin{equation}
\begin{aligned}
p(Y|Z) &= \prod_{i}^{m}p(y_i|Y_{<i},Z)
\end{aligned}
\label{eq:chain_prob}
\end{equation}
where $y_0$ is the "beginning" token (\texttt{<bos>}) of sentence and $Y_{<i}$ is a sequence of previous tokens of the token $y_i$. At an inference time, the decoder generates only one output token and concatenates it with previously generated tokens as additional input to produce a new token. The decoder stops the inference process when it generates the end token (\texttt{<eos>}) or meets the maximum length. 


The model loss $L_{entropy}$ is a cross-entropy loss  that minimizes the sum of negative loglikelihoods of the tokens:
\begin{equation}
\begin{aligned}
L_{entropy} = - \sum^{m}_{j=1}\sum^{}_{w}p_{true}(w|Y_{<j},Z)log( p(w|Y_{<j},Z))
\end{aligned}
\label{eq:entropy_loss}
\end{equation}
where $p_{true}$ is a one-hot distribution which follows:

\begin{equation}
\begin{aligned}
p_{true}(w|Y_{<j},Z) = \bigg\{
\begin{matrix}
1 \;\;\;\;\; w = y_j 
\\
0 \;\;\;\;\; w \neq y_j
\end{matrix}
\end{aligned}
\label{eq:one_hot_dis}
\end{equation}

In the encoder and the decoder, there are six identical layers, each of which has two sub-layers containing a multi-head self-attention mechanism and a simple, position-wise fully connected feed-forward network. Multi-head attention over the encoder's output is used as the third sub-layer of each decoder layer~\cite{vaswani2017attention}.

Scaled Dot-Product Attention and Multi-Head Attention are two attention functions in Transformers. Let's define $Q$ as a query matrix, $K$ as the key matrix, and $Q$ is the value matrix. Let $d_k$, $d_k$ be the dimensions of queries and keys and $d_v$ be the dimension of values. The attention function of Scaled Dot-Product Attention can be calculated as the formula~\cite{vaswani2017attention}:
\begin{equation}
\begin{aligned}
Attention(Q,K,V) = softmax(\frac{QK^T}{\sqrt{d_k}})V
\end{aligned}
\label{eq:attention}
\end{equation}
where the scaling factor $\frac{1}{\sqrt{d_{k}}}$ is used to avoid extremely small gradient of the softmax function when $d_k$ has a large value. 

To allow the model to learn other information from subspace representation in various positions, the Multi-Head Attention is performed with $d_{model}$-dimensional keys, values, and queries~\cite{vaswani2017attention}: 
\begin{equation}
\begin{aligned}
MultiHead(Q,K,V) &= Concat(head_1,head_2,...,head_h)W^O \\
\text{where  } head_i &= Attention(QW_i^Q,KW_i^K,VW_i^V)
\end{aligned}
\label{eq:multihead_att}
\end{equation}
where $h$ is the number of heads while $W^Q_{i} \in \mathbb{R}^{d_{model} \times d_k}$, $W^K_{i} \in \mathbb{R}^{d_{model} \times d_k}$, $W^V_{i} \in \mathbb{R}^{d_{model} \times d_v}$ are the parameter matrices of head $i$. When concatenating these heads, we need to multiply with $W^O$, a parameter matrix belonging to $\mathbb{R}^{hd_v \times d_{model}}$ to get the final multi-head matrix.

\subsubsection{Ranking Models}

Inspired by the works of \citet{zhong2020extractive} and \citet{liu2021simcls}, we create a model for ranking the generated candidates from the best performing models in Phase I. Given a paragraph $X$ and a reference description $\hat{Y}$, the model $\mathcal{G}$  will generate a candidate description \textit{Y = $\mathcal{G}$ (X)}, which is compared to the reference description $\hat{Y}$ by a score $s = M(Y, \hat{Y})$ produced by a metric $M$. We apply an evaluation function $f(.)$, composed of multiple metrics, to produce different similarity scores $s_1, s_2,...,s_n$ between the candidate descriptions ($Y_1$, $Y_2$, ..., $Y_n$) and the source paragraph $X$ based on the given metrics. The similarity score $s_i$ between the candidate description $Y_i$ and the source paragraph $X$ is calculated by:

\begin{equation}
\begin{aligned}
s_i &= f({Y_i}, X) \\
    &= \Bigg\{ 
\begin{matrix}
M_{CosineSim}(Y_i, X) & \text{if using cosine similarity} \\
M_{ROUGE}(Y_i, X) & \text{if using ROUGE}  \\
HM(Y_i, X) & \text{if using harmonic mean}  
\end{matrix}
\end{aligned}
\label{eq:eval_function}
\end{equation}

\begin{equation}
\begin{aligned}
HM(Y_i, X) &= HM(M_{ROUGE}(Y_i, X),M_{CosineSim}(Y_i, X)) \\
 &= \frac{2*M_{ROUGE}(Y_i, X)*M_{CosineSim}(Y_i,  X)}{M_{ROUGE}(Y_i, X) + M_{CosineSim}(Y_i, X)}
\end{aligned}
\label{eq:harmonic_mean}
\end{equation}

\begin{equation}
\begin{aligned}
M_{CosineSim}(Y_i, X) = Cosine(BERT(Y_i), BERT(X))
\end{aligned}
\label{eq:cosine_sim}
\end{equation}

In \Cref{eq:eval_function}, we calculate the score $s_i$ between $Y_i$ and $X$ by different given metrics. If using cosine similarity and ROUGE, $f(.)$ provides cosine similarity and ROUGE values between $Y_i$ and $X$ by corresponding metrics. If using harmonic mean,  $f(.)$ fuses cosine similarity $(M_{CosineSim}$) and ROUGE ($M_{ROUGE}$) using harmonic mean ($HM$) in \Cref{eq:harmonic_mean}. For measuring cosine similarity values, we use the embedding vectors produced by the last hidden layer of BERT as in \Cref{eq:cosine_sim}. In contrast, we use raw texts of $Y_i$ and $X$ for measuring ROUGE values.

Supposedly, cosine similarity and ROUGE scores represent two distinct views of similarity, namely, semantic and lexical, respectively. Thus, fusing them would likely give us a balanced sense of the quality of the candidates. Our intuition shows that a good candidate description should take high and positive cosine similarity values when comparing it to the paragraph. Therefore, for any description with negative values, we set these values to 0 to lower the description in the ranking. In the output, we can have the list of candidates ranked by metrics including the best candidate $\tilde{Y}_b$ which has the highest score:

\begin{equation}
\begin{aligned}
\tilde{Y}_b = argmax_{Y_{i}}(f(Y_{i}, X))
\end{aligned}
\label{eq:argmax}
\end{equation}


As in some works~\cite{Shuyang2021CLIF,chen2020simple,wu2020unsupervised,lee2020contrastive}, ranking models were trained on the difference between positive and negative examples in terms of some metric. The candidate descriptions generated from Phase I can provide a diverse spectrum of data; however, they do not provide negative examples. In this case, contrastive learning reflects the diverse qualities of candidates~\cite{liu2021simcls} more than a contrast between negative and positive examples. Since there are no negative examples, we use a ranking loss based on margin ranking loss to $f(.)$:

\begin{equation}
\begin{aligned}
L_{ranking} = L_{gold} + L_{candidate}
\end{aligned}
\label{eq:ranking_loss}
\end{equation}

\begin{equation}
\begin{aligned}
L_{gold} = \sum_{i}max\Bigl(0,f(X,\tilde{Y}_{i}) - f(X,\hat{Y}) + 	\lambda_{gold}\Bigl)
\end{aligned}
\label{eq:gold_loss}
\end{equation}

\begin{equation}
\begin{aligned}
L_{candidate} = \sum_{i} \sum_{j>i}max\Bigl(0,f(X,\tilde{Y}_{j}) - f(X,\tilde{Y}_{i}) + \lambda_{ij}\Bigl)
\end{aligned}
\label{eq:candidate_loss}
\end{equation}

The ranking loss in \Cref{eq:ranking_loss} is a sum loss of the gold loss and the candidate loss. In the gold loss in \Cref{eq:gold_loss}, we compare the difference between the candidate $\tilde{Y}_i$ and the gold description $\hat{Y}$ against the paragraph $X$. We use hyperparameter \(\lambda\)$_{gold}$ as a gold margin. 
The candidate loss in \Cref{eq:candidate_loss} is a difference between each candidate with the other candidates in their list. At first, the candidate list is sorted in descending order by scores. Then, \textit{\(\lambda\)$_{ij}$ = (j-i)*\(\lambda\)$_{candidate}$} is a margin of each candidate compared to the others~\cite{liu2021simcls}. In the experiment, \textit{\(\lambda\)$_{ij}$ = \(\lambda\)$_{i}$ = i*\(\lambda\)$_{candidate}$} with $i$ is the ranking position, from 1 to list size $n$. The higher ranking a candidate has, the less margin it takes. A default value of 0.01 is set to both \textit{\(\lambda\)$_{gold}$} and \textit{\(\lambda\)$_{candidate}$} in the experiment.

Instead of using the ranking loss $L_{ranking}$, we use validation loss $L_{val}$ to check the model performance.

\begin{equation}
\begin{aligned}
L_{val} = 1 - \frac{1}{N} \sum_{i}^{N} f(\tilde{Y}_{b}, \hat{Y}_{i})
\end{aligned}
\label{eq:val_loss}
\end{equation}

In \Cref{eq:val_loss}, the best candidate $\tilde{Y}_{b}$ is the closest one to the paragraph by metrics or has the highest value $f(\tilde{Y}_{b}, \hat{Y}_{i})$. We count the average value of the best candidates compared to their gold description in the validation set and subtract it from 1 (the gold value) to calculate the validation loss. 

\subsection{Sentiment Consistency}

Wikimedia stresses the importance of keeping neutrality in texts across its projects. Thus, when generating descriptions for Wikidata, we must comply with this principle by guaranting that each machine-generated description will have the same sentiment polarities, especially neutrality as in its paragraph. The sentiment consistency helps to evaluate the quality of machine-generated descriptions in the aspect of sentiment beside the capture of salient information from the paragraph. 

To measure the overall sentiment consistency between the generated description and the input paragraph, we employ Kolmogorov-Smirnov (K-S) test. The Kolmogorov-Smirnov (K-S) test is a test to calculate a distance between two distributions~\cite{massey1951kolmogorov,justel1997multivariate}. Let define $F_{1,N}(x)$ and $F_{2,M}(x)$ are the cumulative distributions with the sample sizes $N$ and $M$ of the first set and second set correspondingly. The distance $D$ between two sets can be calculated by \Cref{eq:kol_smir_distance}, where $sup_x$ is the supremum function.

\begin{equation}
\begin{aligned}
D = sup_{x} |F_{1,N}(x) - F_{2,M}(x)|
\end{aligned}
\label{eq:kol_smir_distance}
\end{equation}

\begin{equation}
\begin{aligned}
c_{\alpha} = \sqrt{-ln(\frac{\alpha}{2}).\frac{1}{2}}
\end{aligned}
\label{eq:alpha_level}
\end{equation}

\begin{equation}
\begin{aligned}
D > c_{\alpha}\sqrt{\frac{N + M}{N \times M}}
\end{aligned}
\label{eq:null_hypothesis_condition}
\end{equation}

The critical value $c_{\alpha}$ at the $\alpha$ level of significance is defined by \Cref{eq:alpha_level}. The null hypothesis (two sets have the same distribution) is rejected at level $\alpha$ if satisfying the condition of the inequality in \Cref{eq:null_hypothesis_condition} for the large numbers of $N$ and $M$.

\section{Experiments}
\subsection{Summarization Evaluation Metrics}
\label{eval_metric}

Currently, there are many metrics used to evaluate the model performance not only for summarization but also for text generation and machine translation tasks. In this section, we focus on several metrics that are generally used for summarization tasks and are broadly mentioned in many scientific papers.

\noindent\emph{ROUGE (Recall-Oriented Understudy for Gisting Evaluation)} is one of the most popular and conventional metrics~\cite{lin2004rouge}. It automatically defines the quality of a summary by measuring an overlap rate between it and the gold summary created by humans. This value is reflected by the number of semantic units (n-gram, word sequences, or word pairs) appearing in both the generated summary and the gold summary. 

\noindent\emph{ROUGE-WE} is an extension of ROUGE that uses Word2Vec embeddings of words taken from summaries to compute the semantic similarity. Therefore, it is more proper for abstractive summarization or substantial paraphrasing on summaries. ROUGE-WE also obtains better correlations with human judgments by Spearman and Kendall coefficients~\cite{ng2015better}.


\noindent\emph{BLEU} is measured by the number of position-independent and overlapped n-grams between generated texts and references with a brevity penalty on the text length. It has several advantages that are execution speed, cheap cost, language independence, and high correlations with human evaluation~\cite{papineni2002bleu}. 

\noindent\emph{METEOR} depends on the matching of unigrams between a hypothesis and a given reference to compute a score for the hypothesis quality by three word-mapping modules: exact, stem, and synonymy~\cite{banerjee2005meteor}. F-mean is calculated as a parametrized harmonic mean of precision P and recall R over single-word matches. METEOR addresses BLEU’s weakness when applied to low-resource languages and has a better correlation with human judgment at the sentence/segment level than BLEU~\cite{lavie2009meteor}.

\noindent\emph{MoverScore} depends on Word Mover’s Distance on contextualized embeddings to compute a semantic distance between summaries and references instead of using traditional semantic units such as words or n-grams for the measures. This metric shows strong generalization capability over many summarization tasks~\cite{zhao2019moverscore}. 

\noindent\emph{BertScore} uses token alignments between generated summaries and references to provide similarity scores. It maximizes cosine similarity between BERT's token embeddings by using a greedy matching~\cite{zhang2019bertscore}.

\noindent\emph{InfoLM} is a recently proposed metric for evaluating summarization and data2text generation tasks. In the family of string-based metrics, ROUGE and BLUE depend on exact matches of semantic units such as n-grams. However, they can not compare two strings based on synonyms. InfoLM overcomes this drawback by using a pre-trained masked language model but does not require training to compute similarity scores between summaries and references over the vocabulary by discrete probability distributions~\cite{colombo2022infolm}.

As most of the reference descriptions in \dataset{} are naturally short and have a high correlation with the source paragraphs, using ROUGE may be enough for evaluating the description quality. However, we still use BertScore, METEOR, and BLEU to have more viewpoints in the result evaluation. Especially, BertScore brings an ability to match paragraphs and descriptions on semantic similarity. ROUGE-N-F-measure, BertScore, METEOR, and BLEU are used to estimate the model performance in two-phase summarization, description generation, and candidate ranking. In another use, ROUGE-1, ROUGE-2, and ROUGE-L precision values are for checking the correlation between paragraphs and descriptions in Section~\ref{section:correlation_para_des}. 

\subsection{Phase I: Description Generation}

\subsubsection{Models}

In Phase I, we use these models:

\begin{itemize}
    \item \textit{Baseline model}: We apply the same baseline for \insplit{} and \exsplit{} on the validation and test sets. As mentioned in \Cref{sec:baseline_method}, the baseline model considers the first element of the instance list as the gold description.
    \item \textit{Pre-trained models}: There are four small-scale pre-trained models, \texttt{BART-base}, \texttt{T5-small}, \texttt{T5-base}, and \texttt{SSR-base} are downloaded from Huggingface to use in the training process.
\end{itemize}

\subsubsection{Experimental Details and Results}
As the trending of summarization tasks, we follow a sequence-to-sequence fashion and apply transfer learning from small-scale pre-trained models to train the data with \texttt{batch\_size=8} and \texttt{epochs=3}. The maximum length of the encoder is 256, while we apply a length of 32 to the decoder to support the content generation of longer descriptions, though their average gold length is $\approx$ 4.5. The function \texttt{Seq2SeqTrainer()} of package \texttt{Transformers} is used for the data training and evaluates validation and test sets by ROUGE-1-F-measure (R-1), ROUGE-2-F-measure (R-2), and ROUGE-L-F-measure (R-L). We let the models automatically adjust the learning rate after each epoch, and save the best model state with the highest metrics on the validation sets.

\begin{table*}[!htb]
\small
\centering
\caption{ROUGE, BertScore (BS), METEOR (ME), and BLEU scores between the generated descriptions and the gold descriptions on the validation set in Phase I. All models use greedy decoding.}
\begin{tabular}{p{2.5cm}p{2cm}llllll}
\hline
\multirow{2}{2.5cm}{Model} & \multirow{2}{2cm}{Topic} & \multicolumn{6}{c}{Validation set} \\
\cline{3-8} & & R-1 & R-2 & R-L & BS & ME & BLEU \\
\hline
BART-base & exclusive & 44.04 & 25.57 & 43.22 & \textbf{89.49} & 36.87 & 7.85 \\
\hline
\textbf{T5-small} & exclusive & \textbf{47.06} & \textbf{26.89} & \textbf{46.09} & 89.41 & \textbf{38.94} & \textbf{8.35} \\
\hline
T5-base & exclusive & 39.52 & 20.68 & 38.78 & 87.44 & 30.83 & 4.46 \\
\hline
SSR-base & exclusive & 27.58 & 14.92 & 25.30 & 86.11 & 35.36 & 3.79 \\
\hline
\textit{Baseline} & exclusive & \textit{36.25} & \textit{16.87} & \textit{35.74} & \textit{87.03} & \textit{25.41} & \textit{2.12} \\
\hline
\hline
\textbf{BART-base} & independent &  \textbf{68.79} & \textbf{53.72} & \textbf{68.34} & \textbf{93.99} & \textbf{62.10} & \textbf{16.54} \\
\hline
T5-small & independent & 64.82 & 48.26 & 64.26 & 93.16 & 58.07 & 14.47 \\
\hline
T5-base & independent & 65.74 & 49.22 & 65.22 & 93.27 & 58.94 & 14.54 \\
\hline
SSR-base & independent & 27.56 & 16.25 & 26.42 & 85.97 & 37.80 &  3.71  \\
\hline
\textit{Baseline} & independent & \textit{20.72} & \textit{8.43} & \textit{20.54} & \textit{83.51} & \textit{13.69} & \textit{0.46} \\
\hline
\multicolumn{7}{c}{* R-1, R-2, and R-L are measured by F-measure values.} \\
\hline
\end{tabular}
\label{tab:generation_val}
\end{table*}

\begin{table*}[!htb]
\small
\centering
\caption{ROUGE, BertScore (BS), METEOR (ME), and BLEU scores between the generated descriptions and the gold descriptions on the test set in Phase I. All models use greedy decoding.}
\begin{tabular}{p{2.5cm}p{2cm}llllll}
\hline
\multirow{2}{2.5cm}{Model} & \multirow{2}{2cm}{Topic} & \multicolumn{6}{c}{Test set} \\
\cline{3-8} & & R-1 & R-2 & R-L & BS & ME & BLEU \\
\hline
BART-base & exclusive & 44.41 & 26.10 & 43.63 & \textbf{89.58}  & 37.20 & 7.99 \\
\hline
\textbf{T5-small} & exclusive & \textbf{46.49} & \textbf{26.20} & \textbf{45.59} & 89.41 & \textbf{38.30} & \textbf{8.03} \\
\hline
T5-base & exclusive & 39.60 & 20.59 & 38.98 & 87.66 & 30.82 & 4.76 \\
\hline
SSR-base & exclusive & 27.43 & 14.76 & 25.23 & 86.09 & 35.12 & 3.67 \\
\hline
\textit{Baseline} & exclusive & \textit{36.25} & \textit{16.91} & \textit{35.73} & \textit{87.09} & \textit{25.44} & \textit{2.21} \\
\hline
\hline
\textbf{BART-base} & independent  & \textbf{69.59} & \textbf{54.59} & \textbf{69.12} & \textbf{94.06} & \textbf{62.82} & \textbf{17.47} \\
\hline
T5-small & independent  & 65.57 & 48.97 & 64.93 & 93.26 & 59.13 & 14.79  \\
\hline
T5-base & independent  & 66.39 & 49.64 & 65.84 & 93.33 & 59.56 & 14.68 \\
\hline
SSR-base & independent & 27.92 & 16.54 & 26.75 & 86.04 & 38.70 & 3.72 \\
\hline
\textit{Baseline} & independent & \textit{20.99} & \textit{8.42} & \textit{20.77} & \textit{83.45} & \textit{13.93} & \textit{0.41} \\
\hline
\multicolumn{7}{c}{* R-1, R-2, and R-L are measured by F-measure values.} \\
\hline
\end{tabular}
\label{tab:generation_test}
\end{table*}

\Cref{tab:generation_val} and \Cref{tab:generation_test} show the results of validation and tests of models with two types of data split: \insplit{} and \exsplit{}. The baseline method considers the first item of each instance's list as the gold description. In \insplit{}, \texttt{T5-small} outperformed the baseline and obtained the best performance. In \exsplit{}, \texttt{BART-base} is the winner. \texttt{SSR-base} models are the worst when they even have lower scores than the baseline in the topic-exclusive split.

We take \texttt{T5-small} and \texttt{BART-base} as the best models in both data splits to generate candidate descriptions for Phase II, candidate ranking. To improve the diversity of the generated descriptions, we set \texttt{num\_beams}, the number of beams of beam search from 1 to 25. When \texttt{num\_beams} equals 1, it means no beam search used.

\subsection{Phase II: Candidate Ranking}

\subsubsection{New datasets}
\label{sec:new_datasets}
The purpose of ranking models is to learn the order of candidates by pre-defined metrics. Therefore, we don't need so many samples for the training process. We extract subsets from training, validation, and test sets of Phase I for creating new sets for the training in Phase II. From a given paragraph, five candidates are generated from the best models (\texttt{T5-small} and \texttt{BART-base}) in Phase I.

The new sets have three components: paragraphs, gold descriptions, and lists of candidates. There are two groups of sets, one for \insplit{} and another for \exsplit{}, which have the same set distribution: the training set (6000 samples, 75\%), the validation set (1000 samples, 12.5\%), and the test set (1000 samples, 12.5\%).

\subsubsection{Models}

In Phase II, we use some models on \textit{new sets} which are mentioned in \Cref{sec:new_datasets}. Note that the results of \texttt{T5-small} and \texttt{BART-base} models in here (\Cref{tab:post_evaluation_val} and \Cref{tab:post_evaluation_test}) are different from those in Phase I (\Cref{tab:generation_val} and \Cref{tab:generation_test}).

\begin{itemize}
    \item \texttt{T5-small} (\insplit{}): the best model from Phase I by \insplit{} on new sets.
    \item \texttt{BART-base} (\exsplit{}): the best model from Phase I by \exsplit{} on new sets.
    \item \texttt{BERT + sim}: the ranking model based on BERT by cosine similarity.
    \item \texttt{BERT + R-1-F}: the ranking model based on BERT by ROUGE-1-F-measure.
    \item \texttt{BERT + sim + R-1-F}: the ranking model based on BERT by fusing cosine similarity and ROUGE-1-F-measure in the form of harmonic mean.
    \item \textit{Gold des. vs. Para.}: The comparison between gold descriptions and paragraphs.
\end{itemize}

\subsubsection{Experiment Details and Results}

\begin{table*}[!htb]
\small
\centering
\caption{ROUGE, BertScore (BS), METEOR (ME), and BLEU scores between the generated descriptions and the paragraphs on the validation set in Phase II. \texttt{T5-small-greedy} and \texttt{BART-base-greedy} are the best models in Phase I, shown in \Cref{tab:generation_val}.}

\resizebox{\textwidth}{!}{
\begin{tabular}{lllllllll}
\hline
\multirow{2}{*}{Phase I: Model}&\multirow{2}{*}{Phase II: Model + Metric} & \multirow{2}{*}{Topic} & \multicolumn{5}{c}{Validation set} \\
\cline{4-9} & & & R-1 & R-2 & R-L &  BS & ME & BLEU \\
\hline
T5-small-greedy & -- & exclusive & 12.71 & 7.13 & 12.27 & 82.18 & 5.19 & 0.55 \\
\hline
T5-small-beam & BERT + sim & exclusive & 24.17 & 19.06 & 23.64 & 84.84 & 13.35 & 5.21 \\
\hline
T5-small-beam & BERT + R-1-F & exclusive & \textbf{25.53} & \textbf{20.29} & \textbf{24.96} & 85.35 & \textbf{14.17}  & \textbf{5.47} \\
\hline
T5-small-beam & BERT + sim + R-1-F & exclusive & \textbf{25.53} & \textbf{20.29} & \textbf{24.96} & \textbf{85.36} & 14.13 & 5.46 \\
\hline
-- & Gold des. vs Para.$^{+}$ & exclusive & 13.55 & 6.85 & 12.58 &  82.60 & 5.57 & 0.43
\\
\hline
\hline
BART-base-greedy & -- & independent & 13.67 &  6.60 & 12.74 & 81.90 & 5.13 & 0.29\\
\hline
BART-base-beam & BERT + sim & independent & 13.97 & 7.20 & 12.88 & 81.81 & 5.33 & 0.61 \\
\hline
BART-base-beam & BERT + R-1-F & independent & \textbf{15.61} & \textbf{8.29} & \textbf{14.45} & 82.38 & 6.09 & 0.63 \\
\hline
BART-base-beam & BERT + sim + R-1-F & independent & \textbf{15.61} & \textbf{8.29} & \textbf{14.45} & \textbf{82.39} & \textbf{6.11} & \textbf{0.64} \\
\hline
-- & Gold des. vs Para.$^{+}$ & independent & 13.03 & 5.61 & 12.03 & 81.69 & 4.78 & 0.19
\\
\hline
\multicolumn{8}{c}{$^{+}$ Gold descriptions against paragraphs. } \\
\hline
\end{tabular}
}
\label{tab:post_evaluation_val}
\end{table*}

\begin{table*}[!htb]
\small
\centering
\caption{ROUGE, BertScore (BS), METEOR (ME), and BLEU scores between the generated descriptions and the paragraphs on the test set in Phase II. \texttt{T5-small-greedy} and \texttt{BART-base-greedy} are the best models in Phase I, shown in \Cref{tab:generation_test}.}

\resizebox{\textwidth}{!}{
\begin{tabular}{lllllllll}
\hline
\multirow{2}{*}{Phase I: Model}&\multirow{2}{*}{Phase II: Model + Metric} & \multirow{2}{*}{Topic} & \multicolumn{5}{c}{Test set} \\
\cline{4-9} & & & R-1 & R-2 & R-L &  BS & ME & BLEU \\
\hline
T5-small-greedy & -- & exclusive & 13.16 & 7.42 & 12.66 & 82.12 & 5.34 & 0.59 \\
\hline
T5-small-beam & BERT + sim & exclusive & 23.93 & 18.50 & 23.30 & 84.71 & 12.71 & 4.56 \\
\hline
T5-small-beam & BERT + R-1-F & exclusive & \textbf{25.36} & 19.61 & 24.64 & \textbf{85.26} & \textbf{13.43} & \textbf{4.77}\\
\hline
T5-small-beam & BERT + sim + R-1-F & exclusive & \textbf{25.36} & \textbf{19.69} & \textbf{24.67} & \textbf{85.26} & \textbf{13.43} & \textbf{4.77}  \\
\hline
-- & Gold des. vs Para.$^{+}$ & exclusive & 13.65 & 6.81 & 12.59 & 82.54 & 5.57 & 0.37
\\
\hline
\hline
BART-base-greedy & -- & independent & 14.06 & 6.74 & 13.04 & 82.09 & 5.38 & 0.16 \\
\hline
BART-base-beam & BERT + sim & independent & 14.47 & 7.41 & 13.18 & 82.11 & 5.69 & 0.48\\
\hline
BART-base-beam & BERT + R-1-F & independent & \textbf{16.35} & \textbf{8.74} & \textbf{15.06} & \textbf{82.67} & \textbf{6.63} &  \textbf{0.51} \\
\hline
BART-base-beam & BERT + sim + R-1-F & independent & \textbf{16.35} & 8.72 & 15.01 & \textbf{82.67} & \textbf{6.63} & \textbf{0.51} \\
\hline
-- & Gold des. vs Para.$^{+}$ & independent & 14.28 & 6.67 & 13.09 & 82.08 & 5.70 & 0.28
\\
\hline
\multicolumn{8}{c}{$^{+}$ Gold descriptions against paragraphs. } \\
\hline
\end{tabular}
}
\label{tab:post_evaluation_test}
\end{table*}

\begin{table*}[!htb]
\small
\centering
\caption{ROUGE, BertScore (BS), METEOR (ME), and BLEU scores between the generated descriptions and the gold descriptions on the validation set in Phase II. \texttt{T5-small-greedy} and \texttt{BART-base-greedy} are the best models in Phase I, shown in \Cref{tab:generation_val}.}


\resizebox{\textwidth}{!}{
\begin{tabular}{lllllllll}
\hline
\multirow{2}{*}{Phase I: Model}&\multirow{2}{*}{Phase II: Model + Metric} & \multirow{2}{*}{Topic} & \multicolumn{5}{c}{Validation set} \\
\cline{4-9} & & & R-1 & R-2 & R-L &  BS & ME & BLEU \\
\hline
T5-small-greedy & -- & exclusive & 40.68 & 21.73 & 39.53 & 89.53 & 33.84 & 6.69 \\
\hline
T5-small-beam & BERT + sim & exclusive & 42.42 & 24.55 & 41.09 & \textbf{89.48} & 39.16 & 9.05 \\
\hline
T5-small-beam & {BERT + R-1-F} & exclusive & \textbf{43.84} & \textbf{26.34} & \textbf{42.48} & 89.41 & \textbf{41.34} & \textbf{9.78}\\
\hline
T5-small-beam & {BERT + sim + R-1-F} & exclusive & 43.63 & 26.02 & 42.27 & 89.39 & 41.07 & 9.75  \\
\hline
\hline
BART-base-greedy & -- & independent & 43.96 & 29.23 & 43.70 & 90.19 & 37.23 & 10.48 \\
\hline
BART-base-beam & BERT + sim & independent & 60.83 & 45.15 & 60.52 & \textbf{93.48} & 54.66 & 16.16\\
\hline
BART-base-beam & BERT + R-1-F & independent & \textbf{66.56} & \textbf{51.44} & \textbf{66.14} & 92.66 & \textbf{61.54} &  \textbf{17.74} \\
\hline
BART-base-beam & {BERT + sim + R-1-F} & independent & 66.44 & 51.25 & 66.02 & 92.65 & 61.35 & 17.54 \\
\hline
\end{tabular}
}
\label{tab:post_evaluation_generated_vs_gold_validation}
\end{table*}

\begin{table*}[!htb]
\small
\centering
\caption{ROUGE, BertScore (BS), METEOR (ME), and BLEU scores between the generated descriptions and the gold descriptions on the test set in Phase II. \texttt{T5-small-greedy} and \texttt{BART-base-greedy} are the best models in Phase I, shown in \Cref{tab:generation_test}.}

\resizebox{\textwidth}{!}{
\begin{tabular}{lllllllll}
\hline
\multirow{2}{*}{Phase I: Model}&\multirow{2}{*}{Phase II: Model + Metric} & \multirow{2}{*}{Topic} & \multicolumn{5}{c}{Test set} \\
\cline{4-9} & & & R-1 & R-2 & R-L &  BS & ME & BLEU \\
\hline
T5-small-greedy & -- & exclusive & 38.26 & 19.94 & 37.27 & 88.97 & 31.77 & 4.70 \\
\hline
T5-small-beam & BERT + sim & exclusive & 42.65 & 25.20 & 41.43 & 89.54 & 39.98 & 8.81 \\
\hline
T5-small-beam & BERT + R-1-F & exclusive & \textbf{44.48} & \textbf{27.11} & \textbf{43.29} & \textbf{89.77} & \textbf{42.59} & \textbf{10.07}\\
\hline
T5-small-beam & BERT + sim + R-1-F & exclusive & 44.37 & 26.88 & 43.18 & \textbf{89.77} & 42.51 & 9.92  \\
\hline
\hline
BART-base-greedy & -- & independent & 55.44 & 40.14 & 55.03 & 92.33 & 47.83 & 10.61 \\
\hline
BART-base-beam & BERT + sim & independent & 59.72 & 45.41 & 59.16 & 93.28 & 53.78 & 16.57 \\
\hline
BART-base-beam & BERT + R-1-F & independent & \textbf{67.79} & \textbf{54.41} & \textbf{67.35} & \textbf{94.46} & \textbf{62.96} &  19.64 \\
\hline
BART-base-beam & BERT + sim + R-1-F & independent & 67.73 & 54.37 & 67.28 & 94.43 & 62.89 & \textbf{19.73} \\
\hline
\end{tabular}
}
\label{tab:post_evaluation_generated_vs_gold_test}
\end{table*}

The best ranking models are saved within 3 epochs by transfer learning from the pre-trained model \texttt{bert-base-cased}. \Cref{tab:post_evaluation_val} and \Cref{tab:post_evaluation_test} show several metrics of the ranking models (Phase II) compared to the best generative models (Phase I) between the generated descriptions and the paragraphs on the validation and test sets. We also compare the gold descriptions versus their paragraphs (\textit{Gold des. vs. Para.}). \texttt{T5-small-greedy} and \texttt{BART-base-greedy} as the best models in Phase I are not better than \textit{Gold des. vs. Para.} when they learn to generate new descriptions. However, their performance is only less than $\approx$ 1 ROUGE in \insplit{} and even surpasses \textit{Gold des. vs. Para.} at least $\approx$ 0.6 ROUGE in the validation set of \exsplit{}. These shreds of evidence 
indicate that the generative models were well-trained. 

\Cref{tab:post_evaluation_generated_vs_gold_validation} and \Cref{tab:post_evaluation_generated_vs_gold_test} show several metrics of the ranking models (Phase II) compared to the best generative models (Phase I) between the generated descriptions and the gold descriptions on the validation and test sets. All the ranking models (Phase II) help to boost the quality of the generated descriptions with the better metric values. In this experiment, the \texttt{BERT + R-1-F} model obtains the best performance while the \texttt{BERT + sim + R-1-F} model is the very close follower.

In general, when applying contrastive learning, the ranking models outperform significantly the generative models in both \insplit{} and \exsplit{}. Between the generated descriptions and the paragraphs, the ranking models obtain from $\approx$ 11 to 13 ROUGE better than the generative models in \insplit{} and from $\approx$ 0.5 to 2 ROUGE in \exsplit{}. Between the generated descriptions and the gold description, the ranking models gain from $\approx$ 1 to 6 ROUGE better than the generative models in \insplit{} and from $\approx$ 4 to 22 ROUGE in \exsplit{}. The best ranking models come from R-1 and the fuse of R-1 and cosine similarity for calculating the rank of the candidates.

\subsection{Sentiment Correlations}

We use a pre-trained model from Huggingface\footnote{\url{https://huggingface.co/cardiffnlp/twitter-xlm-roberta-base-sentiment}} to extract the sentiment polarities --- negative, neutral, and positive --- of the generated descriptions, the paragraphs, and the gold descriptions. The goal here is to measure the sentiment correlations between the generated descriptions and the gold descriptions and between the generated descriptions and the paragraphs. 
\Cref{tab:sentiment_polarity} presents the average values of sentiment polarities over the test sets of the two best methods of Phase II. For each text, we have the distribution formula $P(negative) + P(neutral) + P(positive) = 1$. 

In \dataset{}, most texts are non-opinionated i.e., neutral because they were extracted from Wikimedia projects, which portray neutrality as one of five fundamental principles\footnote{\url{https://en.wikipedia.org/wiki/Wikipedia:Five_pillars}}. This principle represents the stance of Wikimedia in avoiding any bias of content to which millions of users with different viewpoints have contributed. Furthermore, it is also easier for administrators to minimize the harsh arguments on a certain article when forcing the stakeholders toward a neutral point. In this sense, a description should keep the sentiment consistent with the Wikipedia paragraph beside the salient information to reflect the original sentiment, especially the neutrality of that paragraph. This creates a sentiment uniformity of content across Wikimedia projects, including Wikipedia and Wikidata.


\begin{table*}[ht]
\small
\centering
\caption{The average values of sentiment polarities in texts on the test sets of two methods, \texttt{BERT + sim + R-1-F } and \texttt{BERT + R-1-F}.}
\begin{tabular}{p{2cm}lll}
\hline
 & \multirow{2}{2cm}{Polarity} & BERT + sim + R-1-F & BERT + R-1-F \\
 & & (topic-exclusive) & (topic-independent) \\
\hline
\multirow{3}{2.5cm}{Paragraphs} & Negative & 19.53 & 17.30 \\ 
& Neutral & 63.25 & 64.45 \\ 
& Positive & 17.20 & \textbf{18.23} \\ 
\hline
\multirow{3}{4em}{Best descriptions} & Negative & 17.91 & 20.13 \\ 
& Neutral & \textbf{70.05} & 67.12 \\ 
& Positive & 12.03 & 12.73 \\ 
\hline
\multirow{3}{4em}{Gold descriptions} & Negative & 20.45 & \textbf{21.42} \\ 
& Neutral & 67.28 & 65.09 \\ 
& Positive & 12.26 & 13.47 \\ 
\hline
\end{tabular}
\label{tab:sentiment_polarity}
\end{table*}

\begin{figure*}[ht]
\centering
\includegraphics[width=\textwidth]{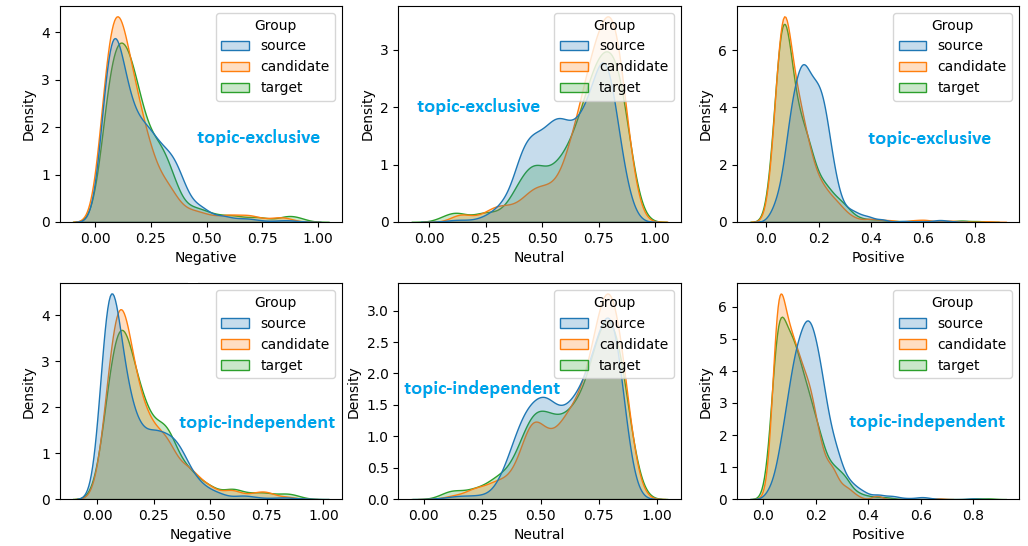}
\caption{The sentiment cumulative distribution by polarities on the test sets produced by two methods, \texttt{BERT + sim + R-1-F} (first row) and \texttt{BERT + R-1-F} (second row). There are 3 sets, the paragraphs (source), the generated descriptions (candidate), and the gold descriptions (target). }
\label{fig:sentiment_density_distribution}
\end{figure*}

\begin{table*}[ht]
\small
\centering
\caption{The critical values ($c_{\alpha }$) and the rejection thresholds ($c_{\alpha}\sqrt{\frac{N + M}{N \times M}}$) by some $\alpha$ levels of significance. Our test sets have the same size of 1000, therefore $N = M = 1000$.}
\begin{tabular}{llllll}
\hline
$\alpha$ & 0.20 & 0.15 & 0.10 & 0.05 & 0.01  \\
\hline
$c_{\alpha}$ & 1.0729 & 1.1380 & 1.2238 & 1.3581 & 1.6276  \\
\hline
$c_{\alpha}\sqrt{\frac{N + M}{N.M}}$ & 0.0479 & 0.0508 & 0.0547 & 0.0607 & 0.0727  \\
\hline

\end{tabular}
\label{tab:alpha_level}
\end{table*}

\begin{figure*}[!htb]
\centering
\includegraphics[width=\textwidth]{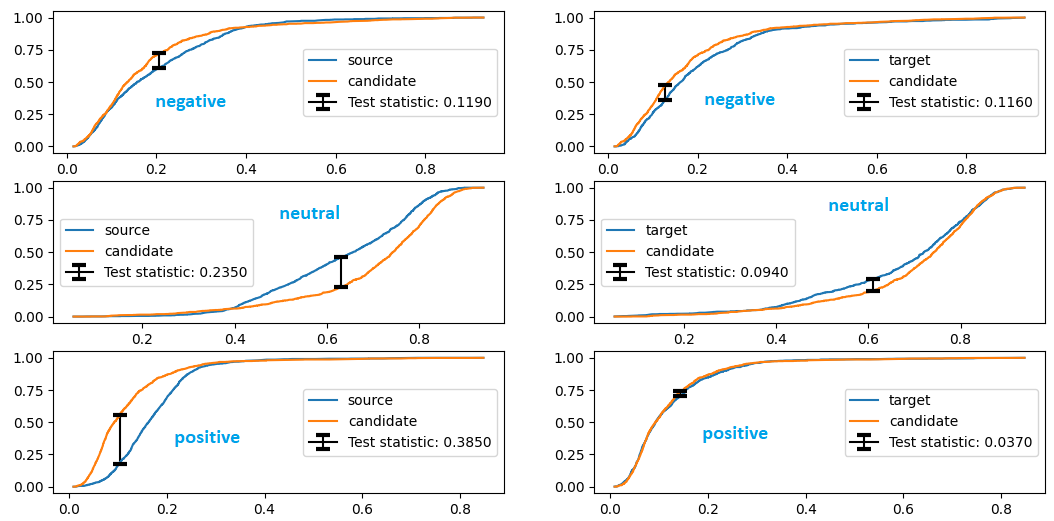}
\caption{The Kolmogorov-Smirnov test on the test sets produced by the method \texttt{BERT + sim + R-1-F} with \insplit{}. The test statistic value is the distance $D$. There are 3 sets, the paragraphs (source), the generated descriptions (candidate), and the gold descriptions (target). }
\label{fig:ks_test1}
\end{figure*}

\begin{figure*}[!htb]
\centering
\includegraphics[width=\textwidth]{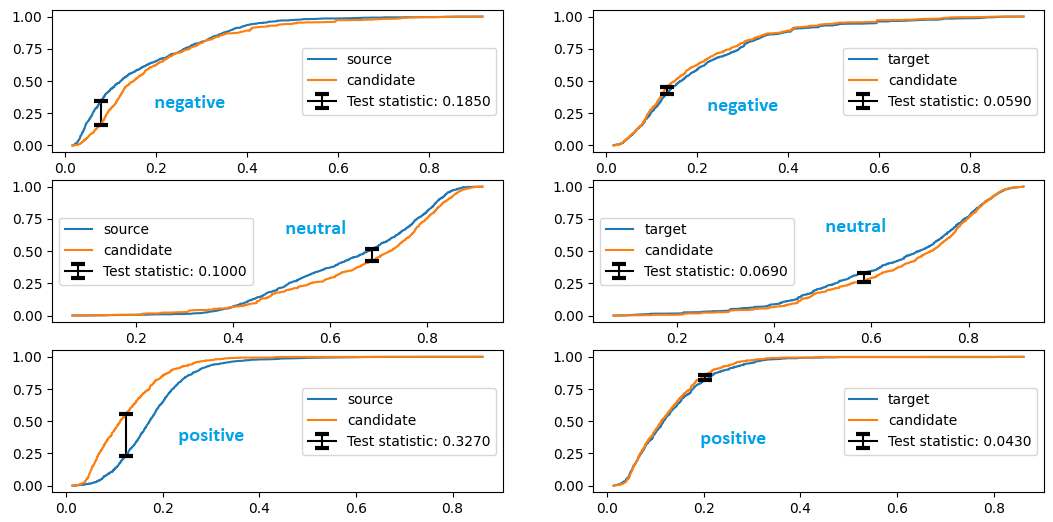}
\caption{The Kolmogorov-Smirnov test on the test sets produced by the method \texttt{BERT + R-1-F} with \exsplit{}. The test statistic value is the distance $D$. There are 3 sets, the paragraphs (source), the generated descriptions (candidate), and the gold descriptions (target). }
\label{fig:ks_test2}
\end{figure*}

The cumulative distributions of semantic polarities by texts are demonstrated in \Cref{fig:sentiment_density_distribution} to provide a general vision of how similar these distributions are. There is no doubt that the distributions of the generated descriptions and the gold descriptions are somewhat analogous in all plots when generation models must learn gold descriptions to create new descriptions. However, the distributions of paragraphs are still independent in some parts, especially in neutral and positive polarities, compared to those of descriptions. This hints that even gold descriptions can not capture all sentiment information in paragraphs.

Although cumulative distributions bring visually a helpful analysis, it is better to obtain a qualitative measure to determine the uniformity of distributions. We perform the Kolmogorov-Smirnov (K-S) test~\cite{massey1951kolmogorov,justel1997multivariate} to compare the distributions of paragraphs vs. generated descriptions and paragraphs vs. gold descriptions. The test output is a statistic value that allows concluding whether two input sets have the same distribution or not. \Cref{tab:alpha_level} lists several critical values by $\alpha$ levels of significance, which are used to validate the condition of \Cref{eq:null_hypothesis_condition} to reject or accept the null hypothesis.

\Cref{fig:ks_test1} and \Cref{fig:ks_test2} show the K-S statistic values on various sets by sentiment polarities. We reject the null hypothesis in all levels of significance between the generated descriptions and the paragraphs. In other say, the distributions of the generated descriptions and the paragraphs are different. Here, the K-S highest distance of positive polarity is up to 0.3850 which indicates that there is a big difference in sentiment distribution between the generated descriptions and the paragraphs. In this case, the generated descriptions can not be considered good descriptions in capturing sentiment information.

In another aspect, the null hypothesis is accepted in some levels of significance between the generated descriptions and the gold descriptions. At the one percent level of significance (0.01), the positive polarity of \texttt{BERT + sim + R-1-F} (\Cref{fig:ks_test1}) and the negative, neutral, and positive polarities of \texttt{BERT + R-1-F} (\Cref{fig:ks_test2}) accept the null hypothesis. At the five percent level of significance (0.05), the null hypothesis is true only with the negative and positive polarities of \texttt{BERT + R-1-F} (\Cref{fig:ks_test2}). Overall, we conclude that training models on \exsplit{} can capture sentiment polarities better than on \insplit{}. Furthermore, the sentiment distributions between the generated description and the gold descriptions are more identical than between the generated descriptions and the paragraphs.

\subsection{Correlation with Human Evaluation}

In this section, we randomly take 100 samples -- 50 samples from the test set from \insplit{} and 50 samples from the test set from \exsplit{} -- for human evaluation. \Cref{tab:human_eval_samples} shows some samples in the human evaluation process, where each sample contains a paragraph, a gold description, and a machine-generated candidate. We label the machine-generated description and the gold description as "Summary 1" and "Summary 2" to avoid bias in the evaluation process. We also include some empty summaries to check the seriousness of evaluators. 

\begin{table*}[!htp]
\small
\centering
\caption{Some typical samples are used for human evaluation. "Machine-generated description" and "Gold description" are the best machine-generated description and the gold description. 
Descriptions with $\checkmark$ were selected by all evaluators.}
\resizebox{\textwidth}{!}{
\begin{tabular}{p{13cm}}
\hline
\# Samples \\
\hline
\textbf{Paragraph 1}: The Berliner Verkehrsbetriebe (German: Berlin Transport Company) is the main public transport company of Berlin, the capital city of Germany. It manages the city's U-Bahn underground railway, tram, bus, replacement services (Ersatzverkehr, EV), and ferry networks, but not the S-Bahn urban rail system. The generally used abbreviation, BVG, has been retained from the company's original name, Berliner Verkehrs-Aktiengesellschaft (Berlin Transportation Stock Company). Subsequently, the company was renamed Berliner Verkehrs-Betriebe. During the division of Berlin, the BVG was split between BVG...\\
\textcolor{blue}{\textbf{Machine-generated description}: public transport company of Berlin, Germany} $\checkmark$ \\ 
\textbf{Gold description}: public transport agency in Berlin  \\
\hline
\textbf{Paragraph 2}: The Harbour Grace CeeBee Stars, (also commonly known as the Harbour Grace Ocean Enterprises CeeBee Stars due to a sponsorship deal that began October 23, 2015) are a senior ice hockey team based in Harbour Grace, Newfoundland and Labrador and part of the Avalon East Senior Hockey League. The CeeBees are eight-time winners of the Herder Memorial Trophy as provincial champions. \\
\textbf{Machine-generated description}: ice hockey team \\
\textcolor{blue}{\textbf{Gold description}: ice hockey team in Harbour Grace, Newfoundland and Labrador} $\checkmark$ \\
\hline
\textbf{Paragraph 3}: The Albanian Urban Lyric Song is a musical tradition of Albania that started in the 18th century and culminated in the 1930s. \\
\textcolor{blue}{\textbf{Machine-generated description}: Albanian musical tradition} $\checkmark$ \\
\textbf{Gold description}: music genre \\
\hline
\textbf{Paragraph 4}: Fuidio is a hamlet and minor local entity located in the municipality of Condado de Treviño, in Burgos province, Castile and León, Spain. As of 2020, it has a population of 26. \\
\textcolor{blue}{\textbf{Machine-generated description}: human settlement in Burgos Province, Castile and León, Spain} $\checkmark$ \\
\textbf{Gold description}: human settlement in Spain \\
\hline
\textbf{Paragraph 5}: On 4 September 2014, 82-year-old Palmira Silva was beheaded in her back garden in Edmonton, London, by 25-year-old Nicholas Salvador, who was on a rampage... Psychiatrists found evidence that Salvador had paranoid schizophrenia. On 23 June 2015, he was found not guilty of murder on the basis of insanity and was detained indefinitely in a psychiatric hospital. \\
\textbf{Machine-generated description}: Psychiatrists found evidence that Nicolas Salvador had paranoid schizophrenia, not guilty of murder on the basis of insanity, and detained indefinitely in a psychiatric hospital  \\
\textcolor{blue}{\textbf{Gold description}: 2014 beheading in Edmonton, London} $\checkmark$ \\
\hline
\textbf{Paragraph 6}: Planar Handbook is an optional supplemental source book for the 3.5 edition of the Dungeons \& Dragons fantasy roleplaying game. \\
\textcolor{blue}{\textbf{Machine-generated description}: supplemental book for the 3.5 edition of the Dungeons \& Dragons fantasy roleplaying game}  $\checkmark$  \\
\textbf{Gold description}: tabletop role-playing game supplement \\
\hline
\textbf{Paragraph 7}: U.S. Games Systems, Inc. (USGS) is a publisher of playing cards, tarot cards, and games located in Stamford, Connecticut. Founded in 1968 by Stuart R. Kaplan, it has published hundreds of different card sets, and about 20 new titles are released annually. The company's product line includes children's card games, museum products, educational cards, motivational cards, tarot cards, and other fortune-telling card decks... \\
\textcolor{blue}{\textbf{Machine-generated description}: American publisher of playing cards, tarot cards, and games}  $\checkmark$  \\
\textbf{Gold description}: card game publishing company \\
\hline
\end{tabular}
}
\label{tab:human_eval_samples}
\end{table*}

\begin{table*}[!htb]
\small
\centering
\caption{Human agreement by evaluation coefficients over 4 criteria. The machine-generated descriptions were obtained from Phase I.}
\resizebox{0.8\textwidth}{!}{
\begin{tabular}{llllll}
\hline
 & Phase I generated. vs gold.  & adequacy & relevance & correctness & concise\\
\hline
$\alpha_{n}$ &  0.7609 & -0.0461 & 0.0793 & 0.2307 & -0.0863 \\
\hline
$\alpha_{i}$ & 0.7609 & -0.0398 & 0.1753 & 0.3175 & -0.0070 \\
\hline
$\kappa_{c}$ & 0.7533 & 0.1123 & 0.1204 & 0.2341 & 0.0627 \\
\hline
$\kappa_{f}$ & 0.7615 & 0.0715 & 0.1137 & 0.2303 & 0.0326 \\
\hline
\textit{S} & 0.8266 & 0.0666 & 0.3911 & 0.6088 & 0.0133 \\
\hline
$\pi$ & 0.7601  & -0.0496 & 0.0762 & 0.2281 & -0.0899 \\
\hline
average score & - & 3.91/5 & 4.57/5 & 4.7/5 & 3.93/5 \\
\hline
distribution & \textbf{23.66\% vs 76.33\%} & -  & - & - & - \\
\hline
\multicolumn{6}{c}{$\alpha_{n}$: alpha nominal, $\alpha_{i}$: alpha interval} \\
\hline
\end{tabular}}
\label{tab:human_evaluation_phase_I}
\end{table*}

\begin{table*}[!htb]
\small
\centering
\caption{Human agreement by evaluation coefficients over 4 criteria. The machine-generated descriptions were obtained from Phase II.}
\resizebox{0.8\textwidth}{!}{
\begin{tabular}{llllll}
\hline
 & Phase II selected. vs gold.  & adequacy & relevance & correctness & concise\\
\hline
$\alpha_{n}$ &  0.3833 & -0.0152 & -0.1742 & -0.0880 & -0.0979 \\
\hline
$\alpha_{i}$ & 0.3833 & 0.0423 & 0.1388 & -0.1191 & -0.2012 \\
\hline
$\kappa_{c}$ & 0.3814 & 0.0370 & -0.0198 & -0.0073 & -0.0063 \\
\hline
$\kappa_{f}$ & 0.3816 & 0.0299 & -0.0190 & -0.0088 & -0.0051 \\
\hline
\textit{S} & 0.3866 & 0.0541 & 0.1000 & 0.3458 & 0.1833 \\
\hline
$\pi$ & 0.3812  & -0.0186 & -0.1781 & -0.0916 & -0.1015 \\
\hline
average score & - & 3.62/5 & 4.31/5 & 4.52/5 & 4.23/5 \\
\hline
distribution & \textbf{45.33\% vs 55.66\%} & -  & - & - & - \\
\hline
\multicolumn{6}{c}{$\alpha_{n}$: alpha nominal, $\alpha_{i}$: alpha interval} \\
\hline
\end{tabular}}
\label{tab:human_evaluation_phase_II}
\end{table*}

We selected three postgraduate students as evaluators. First, we ask them to choose whether the gold description or the best candidate is the accurate description of a given paragraph. Then, for the description they choose, rate several criteria by scores from 1 to 5: bad and can not use = 1,  not recommended for use = 2, fair but need to consider = 3, good = 4, and perfect = 5.

\citet{van2021human} listed at least 17 different criteria gathered from many scientific papers for evaluating the generated texts. Though the authors did not design these criteria in mind for summarization tasks, this work is helpful enough for us to refer to. Following that, we consider using 4 evaluation criteria that are appropriate for short generated descriptions:
\begin{itemize}
    \item \textit{adequacy or informativeness}~\cite{novikova2018rankme}: This criterion is to check whether a generated description captures enough salient information from the paragraph or not.
    \item \textit{relevance or related}: Does a generated description have an appropriate level such as a topic or theme to the paragraph?
    \item \textit{correctness}: This criterion validates the correctness of grammar and facts from a generated description compared to the paragraph.
    \item \textit{concise or brief}: A generated description must be concise enough but not too short, and it still holds the important information from the paragraph. 
\end{itemize}

After the annotation process, we measure the agreement of evaluators, using Krippendorff’s alpha coefficient ($\alpha$)~\cite{klaus1980content}, Cohen's kappa ($\kappa_{c}$)~\cite{cohen1960coefficient}, Fleiss' kappa or multi-kappa ($\kappa_{f}$)~\cite{davies1982measuring}, Bennett, Alpert and Goldstein's S (\textit{S})~\cite{bennett1954communications}, and Scott's Pi ($\pi$)~\cite{scott1955reliability}. They are common statistics used to measure inter-rater reliability between a number of evaluators over categories. Here are the scales of each coefficient:
\begin{itemize}
    \item $\alpha$ and $\kappa_{f}$: The scale range is from $-1$ to $1$, in which $1$ is a perfect agreement, $0$ shows no agreement beyond chance, and negative values indicate disagreement~\cite{zapf2016measuring}.
    \item $\kappa_{c}$: The original author suggested values $\leq$ 0 for no agreement, 0.01–0.20 for none to slight, 0.21–0.40 for fair, 0.41–0.60 for moderate, 0.61–0.80 for substantial, and 0.81–1.00 for almost perfect agreement~\cite{mchugh2012interrater}. $\kappa_{c}$ is formally identical to $\pi$ but has a different calculation of the expected agreement~\cite{craig1981generalization}.
    \item \textit{S}: Scale is a range from $-1$ to $1$. Between the two raters, $1$ indicates perfect agreement, and $-1/(n-1)$ if the proportion of observed agreement $P_o$ equals $0$. The minimum value $-1$ with the number of categories $n$ equals $2$, and it will go toward to $0$ when $n$ increases~\cite{warrens2012effect}.
    \item $\pi$: The value is calculated by formula $\pi = P_o - P_e /(1-P_e)$. $P_o$ is the observed proportion of agreement and $P_e$ is the expected proportion by chance. Scale is a range from a minimum value $-P_e/(1-P_e)$ (when $P_o$ equals $0$) to a maximum value $1$ (when $P_o$ equals $1$)~\cite{craig1981generalization}. 
    
\end{itemize}

To evaluate the descriptions produced by Phase I and Phase II, we consider the following two scenarios:
\begin{itemize}
    \item \textit{Phase I generated vs. gold} (\Cref{tab:human_evaluation_phase_I}): The generated descriptions from Phase I versus the gold descriptions.
    \item \textit{Phase II selected vs. gold} (\Cref{tab:human_evaluation_phase_II}): The selected descriptions from Phase II versus the gold descriptions.
\end{itemize}

\Cref{tab:human_evaluation_phase_I} and \Cref{tab:human_evaluation_phase_II} presents the results of coefficients over criteria with their average values and data distribution on \textit{Phase I generated vs. gold} and \textit{Phase II selected vs. gold} correspondingly. In \Cref{tab:human_evaluation_phase_I}, we have a slight disagreement on criteria \textit{adequacy} and \textit{concise} while we have from a slight to a fair agreement on criteria \textit{relevance} and \textit{correctness}. In \Cref{tab:human_evaluation_phase_II}, we have a slight disagreement on the criteria \textit{adequacy}, \textit{relevance}, \textit{correctness}, and \textit{concise}. 

Meanwhile, there is a high consensus among evaluators in choosing \textit{Phase I generated vs. gold} and \textit{Phase II selected vs. gold}. Especially, there are 45.33\% descriptions chosen for Phase II compared to 23.66\% descriptions chosen for Phase I against the gold descriptions. Therefore, we can conclude that the quality of descriptions in Phase II is better than those in Phase I in human evaluation. Furthermore, it can infer the descriptions taken from Wikidata are not likely qualified as the gold descriptions under the eyes of evaluators. 

When applying contrastive learning, the generated descriptions are closer to the paragraphs because they have a higher semantic similarity. By this, they tend to be longer in length and capture more important information, which impacts the evaluator's decision. In addition, the average scores over the 5-scale of all criteria are generally high, at least 4. Having the lowest value, criterion \textit{adequacy} indicates the missing importance in descriptions. It may be from the difference in expectations for Wikidata descriptions compared to ordinary summaries in abstractive summarization. Wikidata descriptions are generally short and reflect the most prominent information of a Wikipedia paragraph, while an ordinary summary needs to capture all important information, even in the short length.

\subsection{Error Analysis}

The combination of beam search and contrastive learning is a tasty recipe for text summarization, however, there remain two problems in producing quality descriptions. First, we observe that some descriptions contain repetitive texts from setting various beam search configurations in the inference time in Phase I. This makes a false description look closer to its paragraph when measuring the semantic similarity between it and its paragraph in Phase II. Second, it may occur in a case where a few descriptions contain incorrect factual information. The sequence-two-sequence models learn to generate the next tokens with the highest probabilities based on the token frequencies in the dataset. Therefore, a generated description is a sequence of the possible occurrences of tokens instead one with accurate facts. 

\begin{table*}[!htb]

\small
\centering
\caption{A few false descriptions in the output.}
\resizebox{\textwidth}{!}{
\begin{tabular}{p{14cm}}
\hline
\# Errors \\
\hline

\textbf{Paragraph 1}: The Order of Merit of the Federal Republic of Germany (German: Verdienstorden der Bundesrepublik Deutschland, or Bundesverdienstorden, BVO) is the only federal decoration of Germany. It is awarded for special achievements in political, economic, cultural, intellectual or honorary fields... \\
\textbf{Description}: Order of Merit of the Federal Republic of Germany, \hl{awarded for special achievements in political, economic, cultural, intellectual or honorary fields; awarded for special achievements in political, economic, cultural, intellectual or honorary fields...} -- \emph{The description contains repetitive texts.} \\
\hline
\textbf{Paragraph 2}: Malaysian Standard Time (MST; Malay: Waktu Piawai Malaysia, WPM) or Malaysian Time (MYT) is the standard time used in Malaysia. It is 8 hours ahead of Coordinated Universal Time (UTC). The local mean time in Kuala Lumpur was originally GMT+06:46:46... \\
\textbf{Description}: Malaysia \hl{clock clock clock clock clock clock...} -- \emph{The description contains repetitive texts.} \\
\hline
\textbf{Paragraph 3}: Transition metal dinitrogen complexes are coordination compounds that contain transition metals as ion centers the dinitrogen molecules (N2) as ligands. \\
\textbf{Description}: complexes of \hl{protein-coding protein-coding protein-coding protein-coding protein-coding protein-coding protein-coding protein-coding protein-coding...} -- \emph{The description contains repetitive texts.} \\
\hline
\textbf{Paragraph 4}: A shape-up, also called a line-up or an edge-up, is a hairstyle that involves cutting along the natural hairline to straighten it. Shape-ups or edge-ups are the fundamental outline for haircuts today. Edge-ups are typically found among men and short-haired women... \\
\textbf{Description}: a hairstyle that involves cutting along the natural hairline to straighten it. a line-up or an edge-up is \hl{a hairstyle that involves cutting along the natural hairline to straighten it}. -- \emph{Repeat the first sentence.} \\
\hline
\textbf{Paragraph 5}: The National Office of Statistics is the Algerian ministry charged with the collection and publication of statistics related to the economy, population, and society of Algeria at national and local levels. Its head office is in Algiers. \\
\textbf{Description}: \hl{statistical service} -- \emph{The subject is a ministry, not a statistical service.} \\
\hline
\textbf{Paragraph 6}: The Bergen Region is a statistical metropolitan region in the county of Hordaland in Norway. It is centered on the city of Bergen. \\
\textbf{Description}: \hl{geographical feature} -- \emph{The subject is a statistical metropolitan region, not a geographical feature. The description does not contain any token or lemma from the paragraph.} \\
\hline
\textbf{Paragraph 7}: Gargoyle is a free OpenWrt-based Linux distribution for a range of wireless routers based on Broadcom, Atheros, MediaTek and others chipsets, Asus Routers, Netgear, Linksys and TP-Link routers. Among notable features is the ability to limit and monitor bandwidth and set bandwidth caps per specific IP address. \\
\textbf{Description}: \hl{embedded operating system} -- \emph{The subject is a free OpenWrt-based Linux distribution, not a embedded operating system. The description does not contain any token or lemma from the paragraph.} \\
\hline
\end{tabular}
}
\label{tab:poor_description}
\end{table*}

\Cref{tab:poor_description} shows a few false descriptions and their errors, which are highlighted in orange with corresponding explanations. Although our research did not design any mechanism to control the repetitive texts and factual information, these problems were addressed by some approaches such as Pointer-Generator Networks~\cite{See2017GetToThePoint}, Global Encoding~\cite{lin2018global}, reinforcement learning~\cite{zhang2021far}, rule-based/heuristic transformations~\cite{kryscinski2019evaluating,cao2020factual}, and graph attention~\cite{zhu2020enhancing}.




\section{Conclusion}

In this paper, we introduced \dataset{}, a novel summarization dataset with over 80k samples on 6987 topics created by collecting data from Wikipedia and Wikidata. The dataset aims to produce short descriptions from the given paragraphs. In this work, the two-phase summarization --- description generation and candidate ranking --- proved its advantage in boosting the quality of the produced descriptions. The human evaluation confirmed this statement with 45.33\% of descriptions in Phase II chosen compared to 23.66\% of those in Phase I when comparing with the gold descriptions. Furthermore, the difference in sentiment distribution between the descriptions and paragraphs suggests a potential work on integrating sentiment analysis into text summarization.

We performed some analyses on \dataset{} to understand the data distribution, correlations between the paragraphs and the descriptions, and comparison with other existing datasets. In description generation, small-scale pre-trained models were applied to introduce the description diversity by beam search decoding. In candidate ranking, we trained the BERT models to rank the candidate descriptions by various metrics in a contrastive learning setup. Later, we checked the uniformity of sentiment polarity on the descriptions versus the paragraphs to trigger the need of integrating sentiment analysis into text summarization. In addition, we measured the level of evaluators' agreement by several evaluation criteria for the generated descriptions with some popular correlation coefficients. 

Though our work has a limited scope to generate short descriptions from paragraphs, we think it could be extended to other short text generation tasks, e.g. title generation. In the future, we will expand \dataset{} to a multilingual scale and apply other deep-learning methods to eliminate repetitive texts and capture more factual information. Last but not least, the joint modeling of sentiment analysis and text summarization is an interesting direction to improve the quality of summaries. In particular, one could use multitask learning to train the model on two tasks -- text summarization and sentiment analysis -- at the same time. Another potential approach could be to disentangle the sentiment information from the text and propose a distance or similarity measurement loss function to measure sentiment consistency.

\section*{Data Availability Statement}
\newcommand*{\brokenurlwithoutpar}[2]{{\texttt{#1}}\\*{\texttt{#2}}}
The curated dataset is publicly available at:
\begin{itemize}
    \item \url{https://github.com/declare-lab/WikiDes}
\end{itemize}

\section*{Compliance with Ethical Standards}
\begin{itemize}
    \item This article does not contain any studies with human participants or animals performed by any of the authors.
    \item All authors certify that they have no affiliations with or involvement in any organization or entity with any financial interest or non-financial interest in the subject matter or materials discussed in this manuscript.
\end{itemize}

\section*{Acknowledgements}
This research is supported by the Ministry of Education, Singapore, under its AcRF Tier-2 grant (Project no. T2MOE2008, and Grantor reference no. MOE-T2EP20220-0017). Any opinions, findings and conclusions or recommendations expressed in this material are those of the author(s) and do not reflect the views of the Ministry of Education, Singapore.






\bibliographystyle{elsarticle-num-names}
\bibliography{elsarticle-template-num-names}


\end{document}